%% file: egpaper_for_review.tex
 \documentclass[final]{cvpr}

\usepackage{times}
\usepackage{epsfig}
\usepackage{graphicx}
\usepackage{amsmath}
\usepackage{amssymb}
\usepackage{UserCommands}
\usepackage{makecell}
\usepackage{dsfont}
\usepackage{enumitem}
\usepackage{subfigure}
\usepackage{caption}
\usepackage{arydshln}
\usepackage{stmaryrd}
\usepackage[normalem]{ulem}
\usepackage[ruled,vlined]{algorithm2e}
\usepackage[pagebackref=true,breaklinks=true,colorlinks,bookmarks=false]{hyperref}

\def\cvprPaperID{5688} 
\def\confYear{CVPR 2021}

\begin{document}

\title{\Name: Robust Point Cloud Registration using Deep Spatial Consistency}


\author{Xuyang Bai$^{1}$\hspace{0.15cm} Zixin Luo$^{1}$\hspace{0.15cm} Lei Zhou$^{1}$\hspace{0.15cm} Hongkai Chen$^{1}$\hspace{0.15cm} Lei Li $^{1}$\hspace{0.15cm} Zeyu Hu$^{1}$\hspace{0.15cm}  Hongbo Fu$^{2}$\hspace{0.15cm}  Chiew-Lan Tai$^{1}$\ \\
\normalsize $^1$Hong Kong University of Science and Technology \hspace{0.7cm} $^2$City University of Hong Kong \hspace{0.7cm} \normalsize  \\
\tt\small\{xbaiad,zluoag,lzhouai,hchencf,llibb,zhuam,taicl\}@cse.ust.hk \hspace{0.7cm}
\tt\small hongbofu@cityu.edu.hk}

\maketitle

\begin{abstract}

Removing outlier correspondences is one of the critical steps for successful feature-based point cloud registration. 
Despite the increasing popularity of introducing deep learning techniques in this field, spatial consistency, which is essentially established by {a} 
Euclidean transformation between point clouds, has received almost no individual attention in existing learning frameworks. In this paper, we present \Name, a novel deep
neural network that explicitly incorporates spatial consistency for pruning outlier correspondences. 
First, we propose a nonlocal 
feature aggregation module, weighted by both feature and spatial coherence, for feature embedding of the input correspondences. 
Second, we formulate a differentiable spectral matching module, supervised by pairwise spatial compatibility, {to estimate the inlier confidence of each correspondence from the embedded features.}
With modest computation cost, our method outperforms the state-of-the-art hand-crafted and learning-based outlier rejection approaches on several real-world datasets {by a significant margin}. {We also show its wide applicability by combining \Name~with} 
different 3D local descriptors.[\href{https://github.com/XuyangBai/PointDSC/}{code release}]

\end{abstract}

\section{Introduction}
    \input{Introduction.tex}

\section{Related Work}
     \input{RelatedWork.tex}
\begin{figure*}[tbh]
 	\vspace{-0.6cm}
 	\setlength{\abovecaptionskip}{0.1cm}
\centering
        \includegraphics[width=16cm]{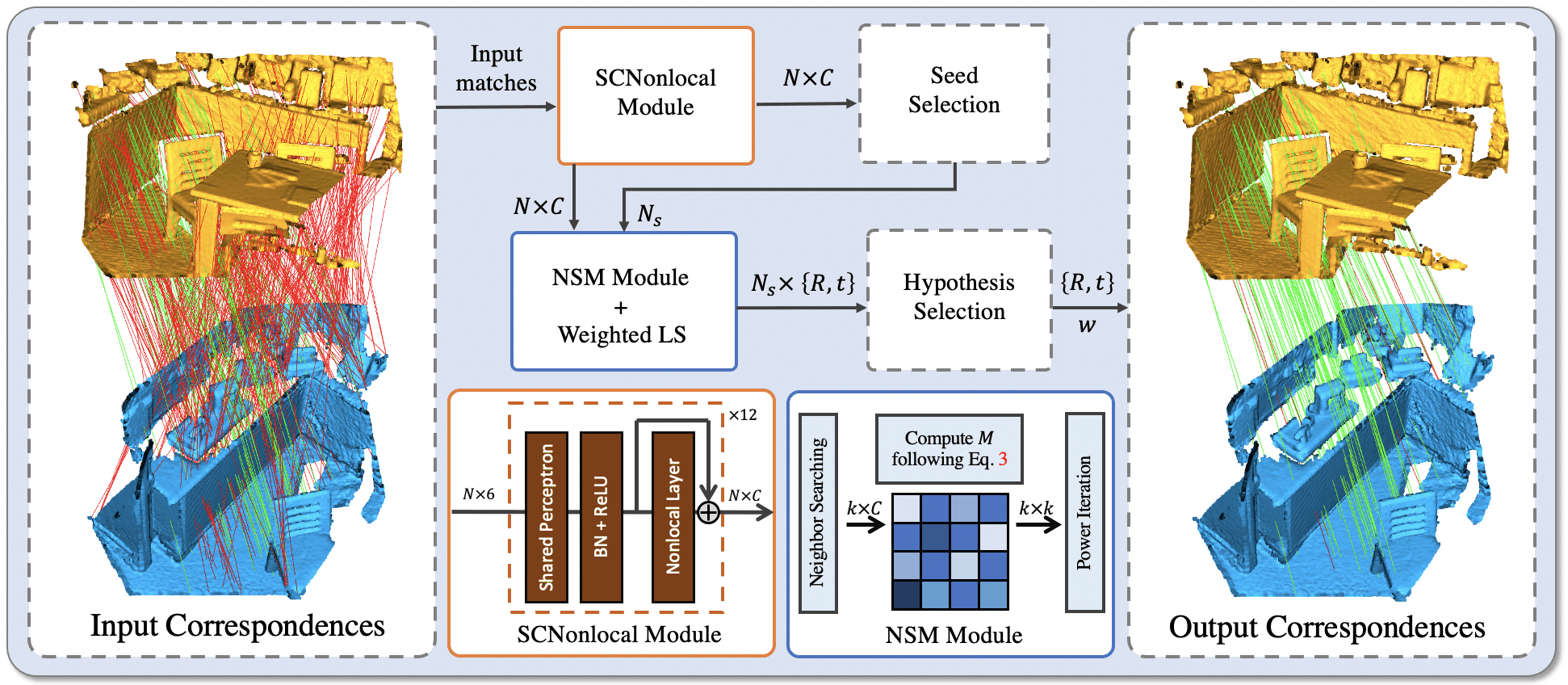}
    \caption{Architecture of the proposed network {\Name}. It takes as input
    the coordinates of putative correspondences, and outputs 
    a rigid transformation and 
    {an inlier/outlier label} 
    for each correspondence. The Spatial Consistency Nonlocal~(\nonlocal) module and the Neural Spectral Matching~(\nsm) module
    are two key components of our network, and 
    perform feature embedding and outlier pruning, respectively. {The green lines and red lines are inliers and outliers, respectively.} LS represents least-squares fitting.
    }
    \label{fig:pipeline}
\end{figure*}

\section{Methodology}
\label{section:methodology}
    \input{Methodology_1.tex}

 \section{Implementation Details}
    \input{Implementation.tex}

\section{Experiments}
  \input{Experiment.tex}

  \input{Experiment_part2.tex}

\section{Conclusion}
    \input{Conclusion.tex}
    
\newpage
{\small
\bibliographystyle{ieee_fullname}
\bibliography{egbib}
}

\clearpage
{ 
\setcounter{page}{1}
\section{Supplementary Material}
  \input{Supplementary.tex}
}

\end{document}

%% file: Introduction.tex
The state-of-the-art feature-based point cloud registration pipelines commonly start from local feature extraction and matching, followed by an outlier rejection for robust alignment. Although 3D local features~\cite{bai2020d3feat, li2020end, Choy_2019_ICCV, gojcic2019perfect, jian20183dfeat} have evolved rapidly, correspondences produced by feature matching are still prone to outliers, especially when the overlap of scene fragments is small. 
In this paper, we focus on developing a robust outlier rejection 
method to mitigate this issue.

Traditional outlier filtering strategies can be broadly classified into two categories, namely the {individual-based} and 
{group-based}~\cite{yang2019performance}. The individual-based approaches, {such as} ratio test~\cite{lowe2004distinctive} and reciprocal check~\cite{opencv_library}, identify inlier correspondences solely based on the descriptor similarity, without considering their spatial coherence. 
In contrast, the group-based methods usually leverage the underlying 2D or 3D scene geometry and identify 
 inlier correspondences {through the analysis of spatial consistency}. 
Specifically, in 2D domain, the spatial consistency \xy{only provides a weak relation between points and epipolar lines}
~\cite{cavalli2020adalam, bian2017gms, zhang2019learning}.
Instead, in 3D domain, spatial consistency is rigorously defined between \textit{every} pair of points by rigid transformations, serving as one of the most important geometric {properties} that inlier correspondences should follow. 
In this paper, we focus on leveraging {the}
spatial consistency in outlier rejection for robust 3D point cloud registration.

\begin{figure}[t]
 \hspace{-0.65cm}
\centering
\setlength{\abovecaptionskip}{0.0cm}
\setlength{\belowcaptionskip}{-0.2cm}
        \includegraphics[width=9cm]{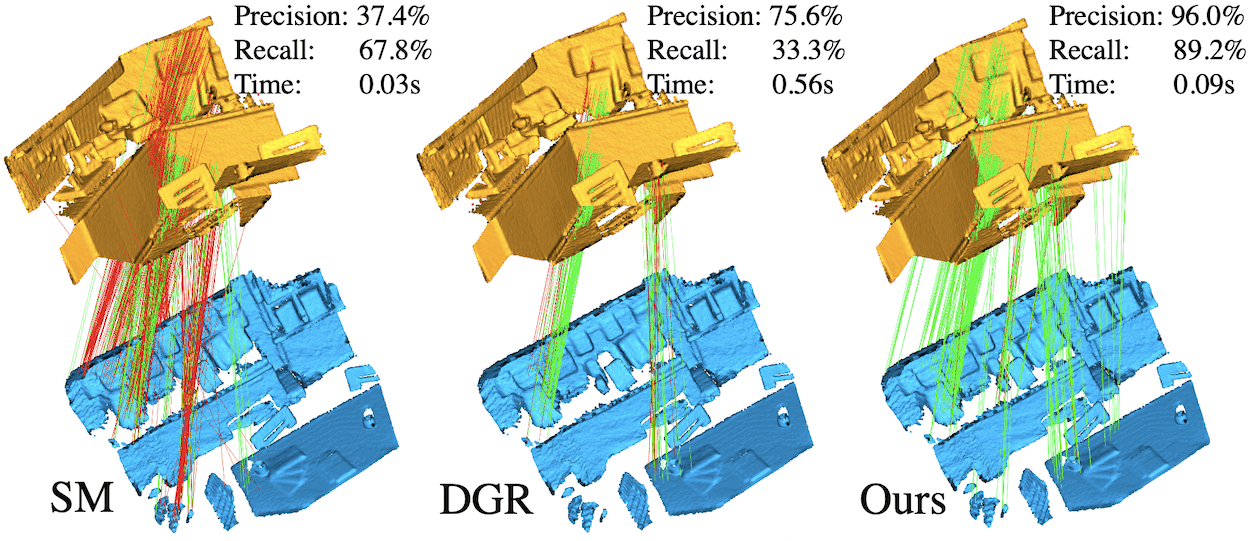}
    \caption{
    Taking advantage of both the superiority of traditional (e.g. SM~\cite{leordeanu2005spectral}) and learning methods (e.g. DGR~\cite{choy2020deep}), our approach integrates important geometric cues into deep neural networks and efficiently identifies inlier correspondences even under high outlier ratios.
    }
    \label{fig:teaser}
\end{figure}

Spectral matching~(SM)~\cite{leordeanu2005spectral} is a
well-known traditional algorithm that heavily relies on 3D spatial consistency for finding inlier correspondences.
It starts with constructing a compatibility graph using the length consistency
{, i.e., preserving the {distance between point pairs}
under rigid transformations,}
then obtains an inlier set by finding the main cluster of the graph through eigen analysis. However, this algorithm has two main drawbacks. 
{First, solely relying on length consistency is intuitive but inadequate because it suffers from the ambiguity problem~\cite{quan2020compatibility}~(Fig.~\ref{fig:ambiguity}a).}
Second, as explained in~\cite{yang2017performance, yang2019performance}, spectral matching cannot effectively handle the case of high outlier ratio~(Fig.~\ref{fig:teaser}, left), 
where {the main inlier clusters become} 
less dominant and thus are difficult to be identified through spectral analysis.


Recently, 
learning-based 3D outlier rejection methods, such as DGR~\cite{choy2020deep} and 3DRegNet~\cite{pais20203dregnet}, formulate outlier rejection as an inlier/outlier classification problem, where the networks embed deep features from correspondence input, and predict {inlier probability} of each correspondence for outlier removal. 
For feature embedding, those methods solely rely on generic operators such as sparse convolution~\cite{choy20194d} and pointwise MLP~\cite{qi2017pointnet} to capture the contextual information, while the essential 3D spatial relations are omitted.
Additionally, 
during outlier pruning, {the} existing methods classify each correspondence only individually,
again overlooking 
the {spatial compatibility} between 
inliers
{and} may hinder 
the classification accuracy.

{All the aforementioned outlier rejection methods are either hand-crafted with spatial consistency adopted, or learning-based without spatial consistency integrated.}
In this paper, we aim to take the best from both line of methods, and propose \Name, a powerful two-stage deep neural network that explicitly leverages the spatial consistency constraints during both feature embedding and outlier pruning.

Specifically, given the point coordinates of input correspondences, we first propose a spatial-consistency guided nonlocal module 
for geometric feature embedding, which captures the relations among different correspondences by {combining the length consistency with feature similarity}
to obtain more representative features.
{Second, we formulate a differentiable spectral matching module, and feed it with not only the point coordinates, but also the embedded features 
to alleviate the ambiguity problem.} 
{Finally, to better handle the small overlap cases, we propose a seeding mechanism, {which} 
first identifies a set of reliable correspondences, then forms several different subsets to perform the neural spectral matching 
multiple times. 
{The best rigid transformation} is finally determined 
such that the geometric consensus is maximized.} 
To summarize, our main contributions are threefold:
\begin{enumerate}[itemsep=-1mm]
\item We propose a spatial-consistency guided nonlocal~(\nonlocal) module for feature embedding, which explicitly leverages the {spatial consistency} to weigh the feature correlation and guide the neighborhood search.
\item We propose a differentiable neural spectral matching~(\nsm) module {based on traditional SM for outlier removal, which goes beyond the simple length consistency metric through deep geometric features.}
\item {Besides showing the superior performance over the state-of-the-arts, our model also demonstrates strong generalization ability from indoor to outdoor scenarios, and wide applicability with different descriptors.}
\end{enumerate}


%% file: RelatedWork.tex
\noindent\textbf{Point cloud registration.} 
Traditional point cloud registration algorithms~(e.g.,~\cite{besl1992method, aiger20084, myronenko2010point, jian2010robust, mellado2014super, mellado2014super}) have been comprehensively reviewed in \cite{pomerleau2015review}.
Recently, learning-based algorithms have been proposed to replace the individual components in the classical registration pipeline, including keypoint detection~\cite{bai2020d3feat, li2019usip, jian20183dfeat} and feature description~\cite{deng2018ppf, deng2018ppfnet, deng20193d, poiesi2020distinctive, bai2020d3feat, Choy_2019_ICCV, gojcic2019perfect, huang2020predator, ao2020SpinNet}. Besides, end-to-end registration networks~\cite{yaoki2019pointnetlk, Wang_2019_ICCV, wang2019prnet, yew2020RPMNet} have been 
proposed.
However, their robustness and applicability in complex scenes cannot always meet expectation,  as observed in~\cite{choy2020deep}, due to highly outlier-contaminated matches.

\noindent\textbf{Traditional outlier rejection.}  RANSAC~\cite{fischler1981random} and its variants~\cite{chum2003locally, barath2018graph, le2019sdrsac, li2020gesac} are still the most popular outlier rejection methods.  However, their major drawbacks are slow convergence and low accuracy in cases with large outlier ratio. 
Such problems become more obvious 
in {3D} point cloud registration since the description ability of 3D descriptors {is} 
generally weaker than those in 2D domain~\cite{lowe2004distinctive, bay2006surf, luo2018geodesc, luo2019contextdesc, luo2020aslfeat} due to the irregular density and the lack of useful texture~\cite{bustos2017guaranteed}. 
Thus, geometric consistency, such as length constraint under rigid transformation, becomes important and is commonly utilized by {traditional} outlier rejection algorithms and analyzed through spectral techniques~\cite{leordeanu2005spectral, cour2007balanced}, voting schemes~\cite{glent2014search, yang2019ranking, sahloul2020accurate}, maximum clique~\cite{perera2012maximal, bustos2019practical, shi2020robin}, random walk~\cite{cho2010reweighted}, belief propagation~\cite{zhou2018learning} or game theory~\cite{rodola2013scale}.
Meanwhile, some algorithms based on BnB~\cite{bustos2017guaranteed} or 
 SDP~\cite{le2019sdrsac} are accurate but usually have {high} time complexity.
Besides, FGR~\cite{zhou2016fast} and TEASER~\cite{yang2019polynomial, Yang20arXivTEASER} are tolerant to outliers from 
robust cost functions such as Geman-McClure function. 
A comprehensive review of traditional 3D outlier rejection methods can be found in~\cite{yang2017performance, yang2019performance}. 

\noindent\textbf{Learning-based outlier rejection.} 
Learning-based outlier rejection methods are first introduced in {the} 2D image matching task~\cite{moo2018learning, zhang2019learning, zhao2019nm, sun2020acne}, where outlier rejection is formulated as an inlier/outlier classification problem. 
{The recent} 
3D outlier rejection methods DGR~\cite{choy2020deep} and 3DRegNet~\cite{pais20203dregnet} follow this idea, and use operators such as sparse convolution~\cite{choy20194d} and pointwise MLP~\cite{qi2017pointnet} to classify the putative correspondences. However, they both ignore the rigid property of 3D Euclidean transformations that has been widely shown to be powerful side information. In contrast, our network explicitly incorporates the spatial consistency between inlier correspondences, constrained by 
rigid transformations, for pruning the outlier correspondences.

%% file: Methodology_1.tex
In this work, we consider two sets of sparse keypoints $\mathbf{X} \in \mathbb{R}^{|\mathbf{X}|\times 3}$ and $\mathbf{Y} \in \mathbb{R}^{|\mathbf{Y}|\times 3}$ from a pair of partially overlapping 3D point clouds, {with each keypoint having} 
an associated local descriptor. 
The input putative correspondence set $C$ can be generated by nearest neighbor search using the local descriptors. Each correspondence ${c_i} \in {C}$ is denoted as ${c_i} = (\bm{x_i, y_{i}}) \in \mathbb{R}^6$, where $\bm{x_i}\in \mathbf{X}, \bm{y_{i}}\in \mathbf{Y}$ are the coordinates of {a pair of 3D keypoints from the two sets.}
{Our objective is to find an inlier/outlier label for $c_i$, being $w_i=1$ and $0$, respectively,}
and recover {an optimal 3D rigid} 
transformation $\mathbf{\hat R},\mathbf{\hat t}$ between the two point sets.
The pipeline of our network~\Name~is shown in Fig.~\ref{fig:pipeline} and can be summarized as follows:
\begin{enumerate}[itemsep=-1mm]
    \item We embed the input correspondences into high dimensional geometric features using the \nonlocal~module
  (Sec.~\ref{subsec:nonlocal}).
    \item We estimate the initial confidence $v_i$ of each correspondence $c_i$ to select a limited number of highly confident and well-distributed \textit{seeds} (Sec.~\ref{subsec:seed}).
	\item For each \textit{seed}, we search for its $k$ nearest neighbors in the feature space and perform neural spectral matching~(NSM) to obtain {its} confidence {of} being an inlier. The confidence values are used 
	to {weigh} 
	the least-squares fitting for computing a rigid transformation for each seed (Sec.~\ref{subsec:nsm}). 
	
	\item The best transformation matrix is selected from all the hypotheses as the one that maximizes the number of inlier correspondences (Sec.~\ref{subsec:hypo_select}).
\end{enumerate}

\subsection{\Name~vs. RANSAC}
\label{sec:vs_ransac}

Here, we clarify the difference between \Name~and RANSAC {to help understand the insights behind our algorithm}.
{Despite} not being designed for improving
classic RANSAC, our \Name~shares {a} 
\textit{hypothesize-and-verify} pipeline {similar to RANSAC}. 
In the sampling step, instead of randomly sampling minimal subsets iteratively, we utilize the learned embedding space to retrieve a pool of larger correspondence subsets in one shot~(Sec.~\ref{subsec:nonlocal} and Sec.~\ref{subsec:seed}). {The correspondences in such subsets} 
have higher probabilities {of being} inliers {thanks to the highly confident seeds and the discriminative embedding space}. In the model fitting step, our neural spectral matching module~(Sec.~\ref{subsec:nsm}) effectively prunes the potential outliers in the retrieved subsets, producing a correct model even when starting from a not-all-inlier sample. In this way, \Name~can tolerate large outlier ratios and produce highly precise registration results, without needing exhaustive iterations. 

\subsection{Geometric Feature Embedding}
\label{subsec:nonlocal}

The first module of our network is the \nonlocal~module, which receives the correspondences $C$ as input and produces a geometric feature for each correspondence. Previous networks~\cite{choy2020deep, pais20203dregnet} learn the feature embedding through generic operators,
 ignoring the unique properties of 3D rigid transformations.
Instead, our \nonlocal~module explicitly utilizes the spatial consistency between inlier correspondences to learn a discriminative 
{embedding space, where inlier correspondences are close to each other.} 

\begin{figure}[t]
\setlength{\abovecaptionskip}{0.05cm}
\setlength{\belowcaptionskip}{-0.35cm}
	\vspace{-0.4cm}
	\centering
    \includegraphics[width=5.5cm]{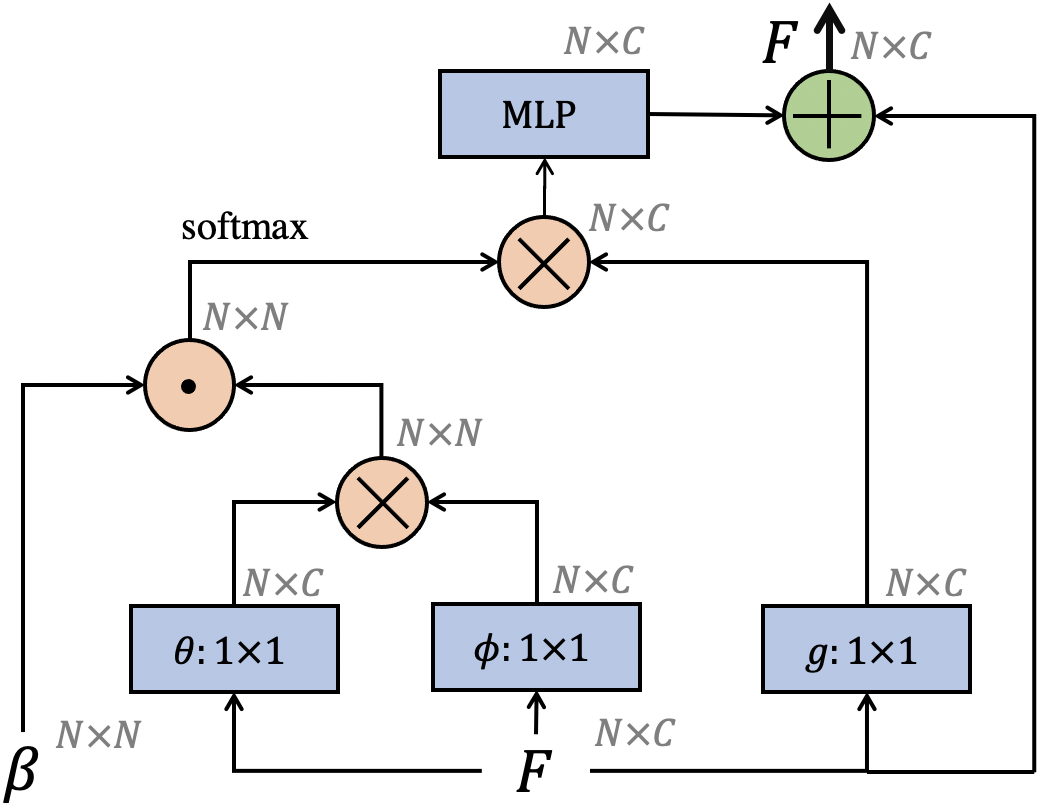}
    \caption{The spatial-consistency guided nonlocal layer. $\bm{\beta}$ represents the spatial consistency matrix calculated using Eq.~\ref{eq:spatial_consistency} and $\bm{F}$ is the feature from {the} previous layer. 
    }
    \label{fig:nonlocal}
\end{figure}

\begin{figure}[tb]
\setlength{\abovecaptionskip}{0.2cm}
\setlength{\belowcaptionskip}{-0.35cm}
	\vspace{-0.5cm}
    \includegraphics[width=8.5cm]{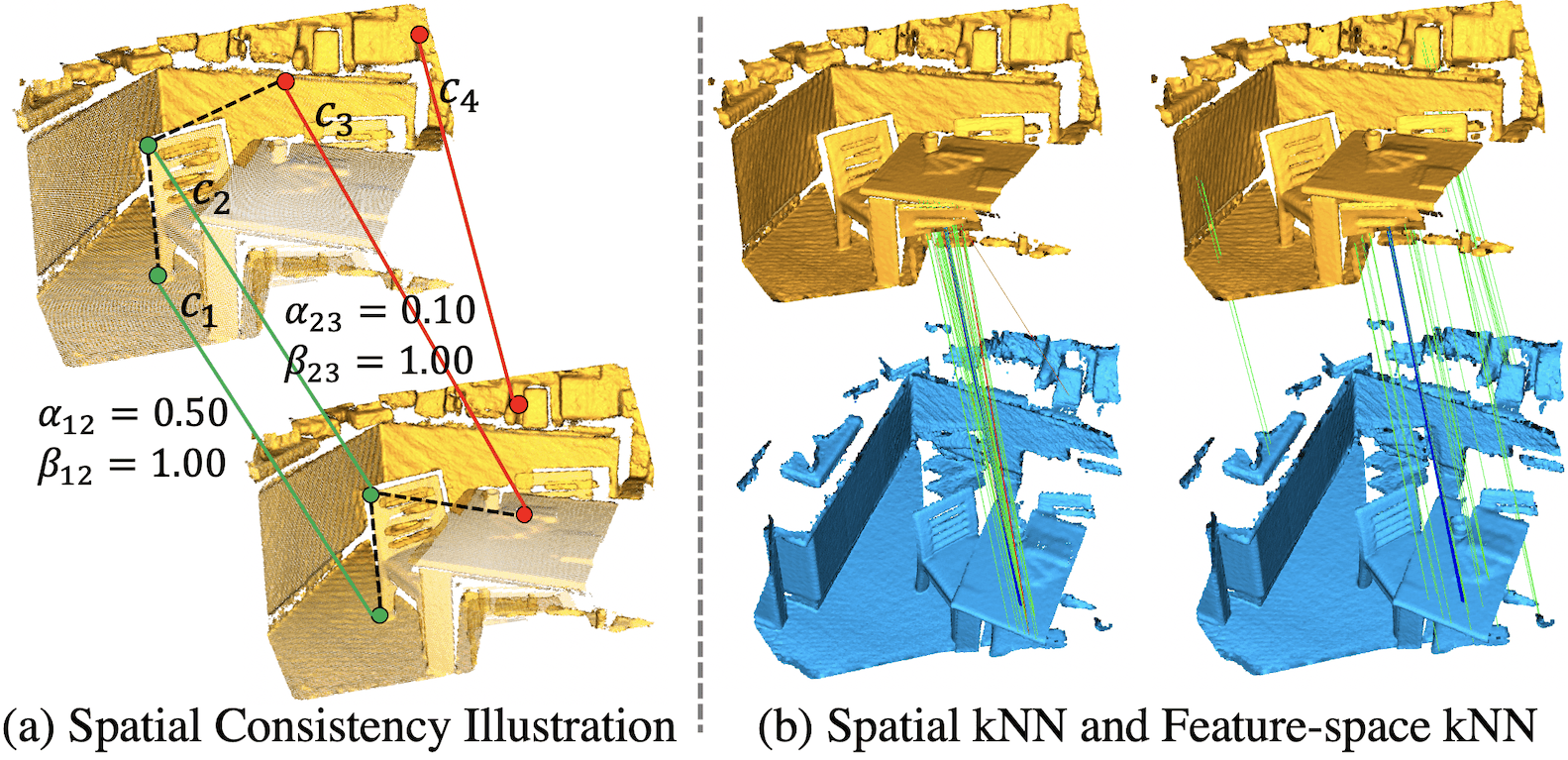}
    \caption{(a) 
    {Inlier correspondence pairs~($c_1,c_2$) always satisfy the length consistency, while outliers~(e.g.~$c_4$) {are usually not spatially consistent with} either inliers ($c_1,c_2$) or {other} outliers~(e.g.~$c_3$).
    However, there exist ambiguity when inliers~($c_2$) and outliers~($c_3$) {happen to} satisfy the length consistency.}
    The feature similarity term $\bm{\alpha}$ provides the possibility to alleviate {the ambiguity issue}. 
    (b) 
    The correspondence subsets of {a seed (blue line)}
     found by spatial kNN~(Left) and feature-space
      kNN~(Right). } 
    \label{fig:ambiguity}
\end{figure}

{As illustrated in Fig.~\ref{fig:pipeline}}, our \nonlocal~module has 12 blocks, each of which consists of 
a shared Perceptron layer, a BatchNorm layer with ReLU, and the proposed nonlocal layer. 
Fig.~\ref{fig:nonlocal} {illustrates this new nonlocal layer}. Let $\bm{f_i} \in \bm{F}$ be the intermediate {feature} representation for correspondence $c_i$. {The design of our nonlocal layer {for updating the features} draws inspiration from the well-known nonlocal network~\cite{wang2018non}, {which captures the long-range dependencies using nonlocal operators.}
Our contribution is to introduce a novel 
{spatial consistency term to complement the feature similarity in nonlocal operators.}
Specifically, we update the features using the following equation:  
}
\begin{equation}
\setlength\abovedisplayskip{3.5pt}\setlength\belowdisplayskip{3.5pt}
	\bm{f_i} = \bm{f_i} + \text{MLP}(\sum\nolimits_{j}^{|C|}\text{softmax}_j(\bm{\alpha\beta}) g(\bm{f_j})) ~,
	\label{eq:nonlocal}
\end{equation}
where $g$ is a linear projection function. 
The feature similarity term $\bm{\alpha}$ is defined as the embedded dot-product similarity~\cite{wang2018non}.
The {spatial consistency} term $\bm{\beta}$ {is defined based on} the {length constraint} of 3D rigid transformations, 
as illustrated in Fig.~\ref{fig:ambiguity}a~($c_1$ and $c_2$). 

Specifically, we {compute} $\bm{\beta}$ by measuring the {length difference} {between the line segments of point pairs in $\mathbf{X}$ and its corresponding segments in $\mathbf{Y}$:} 
\begin{equation}
\setlength\abovedisplayskip{2pt}
\setlength\belowdisplayskip{2.5pt}
    \beta_{ij} = [1 - \frac{d_{ij}^2}{\sigma_d^2}]_+, ~~
    d_{ij} = \big | \left\Vert \bm{x_i-x_j}\right\Vert - \left\Vert \bm{y_i-y_j}\right\Vert \big |,
    \label{eq:spatial_consistency}
\end{equation}
where $[\cdot]_+$ is the $\max (\cdot, 0)$ operation to ensure {a non-negative value of $\beta_{ij}$}, 
and $\sigma_d$ is a {distance} parameter~(see Sec.~\ref{sec:imp}) 
to control the sensitivity to the length difference. Correspondence pairs having the length difference larger than $\sigma_d$ are considered to be incompatible and get zero for $\bm{\beta}$. 
In contrast, ${\beta}_{ij}$ gives a large value only if the two correspondences {$c_i$ and $c_j$} are spatially compatible, serving as a reliable regulator to the feature similarity term. 

Note 
that other forms of spatial consistency can also be easily incorporated here. However, taking 
{an angle-based spatial consistency} 
constraint as an example, the normals of input keypoints might not {always} be available {for} 
outlier rejection and the normal estimation {task is challenging on its own} 
especially for LiDAR point clouds~\cite{zhao2019robust}.
Our \nonlocal~module produces {for each correspondence $c_i$} a feature representation $\bm{f_i}$, which will be used in both seed selection and neural spectral matching module.

\subsection{Seed Selection}
\label{subsec:seed}
{As mentioned before, the traditional spectral matching {technique} has {difficulties} 
in finding {a} 
dominant inlier cluster in low overlapping cases, {where it would fail} 
to provide a clear separation between inliers and outliers~\cite{yang2019extreme}. {In such cases}, directly using the output from spectral matching in weighted least-squares fitting~\cite{besl1992method} for transformation estimation may lead to a sub-optimal solution since
there are still many outliers not being explicitly rejected.} 
To address this issue, inspired by~\cite{cavalli2020adalam}, we design a seeding mechanism to apply neural spectral matching locally. We first find reliable and well-distributed correspondences as seeds, {and around them search for consistent correspondences in the feature space.} 
Then each subset is expected to have a higher inlier ratio {than the input correspondence set}, {and is thus} 
easier for neural spectral matching to find {a} 
correct cluster.  

To select the seeds, we first adopt an MLP to estimate the initial confidence $v_i$ of each correspondence using the feature $\bm{f_i}$ learned by the \nonlocal~module,
and then apply Non-Maximum Suppression~\cite{lowe2004distinctive} over the confidence to find the well-distributed seeds.
{The selected seeds will be used to form multiple correspondence subsets for the neural spectral matching.}

\subsection{Neural Spectral Matching}
\label{subsec:nsm}
In this step, we leverage the learned feature space to augment each seed with {a subset of} consistent correspondences by performing 
$k$-nearest neighbor searching in the feature space.  
We then adopt {the proposed }neural spectral matching~(NSM)
over each subset to estimate {a} 
transformation as one hypothesis. 
Feature-space kNN has several advantages over spatial kNN, as illustrated in Fig.~\ref{fig:ambiguity}b. 
First, {the} neighbors found in the feature space are more likely to follow a similar transformation as the seeds, thanks to the \nonlocal~module. Second, the neighbors chosen in {the} feature space can be located far apart in {the} 3D space,
leading to more robust {transformation} estimation results. 

\begin{figure}[tb]
\setlength{\belowcaptionskip}{-0.35cm}
	\centering
	\vspace{-0.5cm}
    \includegraphics[width=6cm]{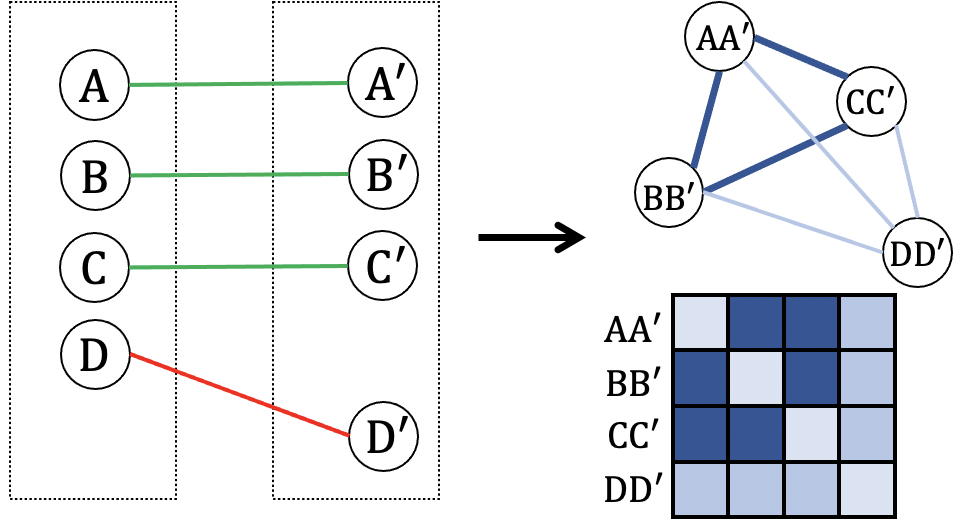}
    \caption{Constructing the compatibility graph {and associated matrix~(Right)} from the input correspondences~(Left). We set the matrix diagonal to zero following~\cite{leordeanu2005spectral}.
    The weight of each graph edge 
    represents the pairwise compatibility between two {associated} correspondences. } 
    \label{fig:assignment_graph}
\end{figure}

Given the correspondence subset $C' \subseteq C~(|C'|=k)$ of each seed constructed by kNN search, we apply NSM to estimate the inlier probability, which is subsequently used in the weighted least-squares
 fitting~\cite{besl1992method} for transformation estimation.  Following~\cite{leordeanu2005spectral}, we first construct a matrix $\mathbf{M}$ representing a compatibility graph associated with $C'$, 
as illustrated in Fig.~\ref{fig:assignment_graph}. Instead of solely relying on the length consistency as~\cite{leordeanu2005spectral}, {we further incorporate the geometric feature similarity to tackle the ambiguity problem}
as illustrated in Fig.~\ref{fig:ambiguity}a. 
Each entry ${M_{ij}}$ measures the compatibility between correspondence $c_i$ and $c_j$ from $C'$, which is defined as 
\begin{equation}
\setlength\abovedisplayskip{5pt}
\setlength\belowdisplayskip{5pt}
    M_{ij} = \beta_{ij} * \gamma_{ij},
	\label{eq:compatibility}
\end{equation}
\begin{equation}
	 \gamma_{ij} = [1 - \frac{1}{\sigma_f^2}  \left\Vert \bm{\bar f_i - \bar f_j} \right\Vert^2]_+
     \label{eq:feat_sm}
\end{equation}
where $\beta_{ij}$ is {the same as} in Eq.~\ref{eq:spatial_consistency},
 $\bm{\bar f_i}$ and $\bm{\bar f_j}$ are the L2-normalized feature vectors, and $\sigma_f$ is a parameter to control sensitivity to {feature difference~(see Sec.~\ref{sec:imp}).} 

 The elements of $\mathbf{M}$ defined above {are} 
 always non-negative and increase with the compatibility between correspondences. Following~\cite{leordeanu2005spectral}, we consider the leading eigenvector of matrix $\mathbf{M}$ as the \textit{association} of each correspondence with {a} 
 main cluster. Since this main cluster is statistically formed by the inlier correspondences, it is natural to interpret this \textit{association} as the inlier probability. The higher the association to the main cluster, the higher the probability of {a correspondence} being an inlier.
The leading eigenvector $\bm{e} \in \mathbb{R}^k$ can be efficiently computed by the power iteration algorithm~\cite{mises1929praktische}. 
We regard $\bm{e}$  
as the inlier probability, since only the relative value of $\bm{e}$ matters.
{Finally} 
we use the probability $\bm{e}$ as the weight to estimate the transformation through {least-squares fitting},
\begin{equation}
\setlength\abovedisplayskip{5pt}\setlength\belowdisplayskip{5pt}
\mathbf{R'},\mathbf{t'} = \arg\min_{\mathbf{R}, \mathbf{t}} \sum\nolimits_{i}^{|C'|} \bm{e_i} \left\Vert \mathbf{R}\bm{x_i} + \mathbf{t} - \bm{y_i}  \right\Vert^2.
    \label{eq:solve_r_t}
\end{equation}
Eq.~\ref{eq:solve_r_t} can be solved in closed form by SVD~\cite{besl1992method}. For the sake of completeness, {we provide} its derivation 
in the supplementary~\ref{supp:proof}. By 
{performing}
such steps for each seed {in parallel}, the network produces a set of transformations $\{\mathbf{R'},\mathbf{t'}\}$ for hypothesis selection.

\subsection{Hypothesis Selection}
\label{subsec:hypo_select}
The final stage of \Name~involves selecting the best hypothesis among the transformations produced by the NSM module. 
The criterion for selecting the best transformation is based on the number of correspondences satisfied by {each}
transformation, 
\begin{equation}
\setlength\abovedisplayskip{5pt}\setlength\belowdisplayskip{5pt}
    \mathbf{\hat R},\mathbf{\hat t} = \arg\max_{\mathbf{R', t'}} \sum\nolimits_{i}^{|C|}  \big \llbracket  \vert\vert \mathbf{R'}\bm{x_i} + \mathbf{t'} - \bm{y_i} \vert\vert < \tau \big \rrbracket,
\end{equation}
where $\llbracket \cdot \rrbracket$ is the Iverson bracket and $\tau$ denotes {an} 
inlier threshold. The final inlier/outlier labels $\bm{w}\in \mathbb{R}^{|C|}$ {are} 
given by ${w_i} = \llbracket \vert\vert \mathbf{\hat R}\bm{x_i} + \mathbf{\hat t} - \bm{y_i} \vert\vert< \tau \big \rrbracket$. 
We then 
recompute the transformation matrix using all the surviving inliers in a least-squares manner, which is a common practice~\cite{chum2003locally, barath2018graph}. 

\subsection{Loss Formulation}

{Considering the compatibility graph illustrated in Fig.~\ref{fig:assignment_graph}, previous works~\cite{choy2020deep, pais20203dregnet} mainly adopt node-wise losses, which supervise each correspondence individually. In our work, we further design an edge-wise loss to supervise the pairwise relations between the correspondences.}

\noindent\textbf{Node-wise supervision.} We denote $\bm{w}^* \in \mathbb{R}^{|C|}$ as the ground-truth inlier/outlier labels constructed by 
\begin{equation}
\setlength\abovedisplayskip{5pt}\setlength\belowdisplayskip{5pt}
	{w}^*_i = \llbracket \vert\vert \mathbf{R^*}\bm{x_i} + \mathbf{t^*} - \bm{y_i} \vert\vert < \tau \big \rrbracket,
\end{equation} where $\mathbf{R^*}$ {and} $\mathbf{t^*}$ are the ground{-}truth rotation and translation matries, respectively.
Similar to~\cite{choy2020deep,pais20203dregnet}, we first adopt the binary cross entropy loss as the node-wise supervision for learning the initial confidence by 
\begin{equation}
\setlength\abovedisplayskip{4pt}\setlength\belowdisplayskip{4pt}
 	L_{class} = \text{BCE}(\bm{v, w^*}),
\end{equation}
where $\bm{v}$ is the initial confidence predicted (Sec.~\ref{subsec:seed}).


\noindent\textbf{Edge-wise supervision} 
We further propose the spectral matching loss as our edge-wise supervision, formulated as
\begin{equation}
\setlength\abovedisplayskip{5pt}\setlength\belowdisplayskip{5pt}
    L_{sm} = \frac{1}{|C|^2}\sum\nolimits_{ij} (\gamma_{ij} - \gamma^*_{ij})^2, 
    \label{eq:sm_loss}
\end{equation} 
where $\gamma^*_{ij}=\llbracket c_i, c_j \text{~are both inliers}\rrbracket$ is the ground-truth compatibility value and $\gamma_{ij}$ is the estimated compatibility value based on the feature similarity defined in Eq.~\ref{eq:feat_sm}.
This loss supervises the relationship between each pair of correspondences, serving as a complement to the node-wise supervision. Our experiments~(Sec.~\ref{subsec:ablation}) show that the proposed $L_{sm}$ remarkably improves the performance.


The final loss is a weighted sum of the two losses,
\begin{equation}
\setlength\abovedisplayskip{5pt}\setlength\belowdisplayskip{5pt}
	L_{total} = L_{sm} + \lambda L_{class},
	\label{eq:full_loss}
\end{equation}
where $\lambda$ is a hyper-parameter to balance the two losses.

%% file: Implementation.tex
\label{sec:imp}
\noindent \textbf{Training.} {We implement our network} 
in PyTorch~\cite{paszke2017automatic}. 
Since each pair of point clouds may have different numbers of correspondences, we randomly sample 1,000 correspondences from each pair to build the batched input during training and set the batch size to 16
point cloud pairs. 
For NSM, we choose the neighborhood size to be $k=40$. (The choice of $k$ is studied in the supplementary~\ref{supp:exp}). 
 {We make $\sigma_f$ learned by the network, and set $\sigma_d$ as 10cm for indoor scenes and 60cm for outdoor scenes,
since $\sigma_d$ has {a} clear physical meaning~\cite{leordeanu2005spectral}.}
 The  hyper-parameter $\lambda$ is set to 3. We optimize the network using the ADAM optimizer with an initial learning rate of 0.0001 and an exponentially decayed factor of 0.99, and train the network for 100 epochs. All the experiments are conducted on a single RTX2080 Ti graphics card.

\noindent \textbf{Testing.} During testing, we use {a} 
full correspondence set as input. We adopt Non-Maximum Suppression~(NMS) to ensure spatial uniformity of the selected seeds, and set the radius for NMS to be the same value as the inlier threshold $\tau$. To avoid {having} excessive seeds {returned} by NMS and make the computation cost manageable, we keep at most 10\% of the input correspondences as seeds.
To improve the precision of the {final} transformation matrix, 
we further adopt a simple yet effective post-refinement stage analogous to iterative re-weighted least-squares~\cite{holland1977robust, bergstrom2014robust}.
The detailed algorithm can be found in the supplementary~\ref{supp:post}.

%% file: Experiment.tex
The following sections are organized as follows. First, we evaluate our method~(\Name) in pairwise registration tasks on 3DMatch dataset~\cite{zeng20173dmatch} (indoor settings) with different descriptors, including the learned ones and hand-crafted ones, in Sec.~\ref{subsec:3dmatch}.  Next, we study the generalization ability of \Name~on KITTI dataset~\cite{Geiger2013IJRR} (outdoor settings) using the model trained on 3DMatch in Sec.~\ref{subsec:kitti}. We further evaluate \Name~in multiway registration tasks on augmented ICL-NUIM~\cite{choi2015robust} dataset in Sec.~\ref{subsec:multi}.
Finally, we conduct ablation studies to demonstrate the importance of each proposed component in \Name.

\subsection{Pairwise Registration}
\label{subsec:3dmatch}

We follow the same evaluation protocols in 3DMatch to prepare  training and testing data, where the test set contains eight scenes with $1,623$ partially overlapped point cloud fragments and their corresponding transformation matrices. We first voxel-downsample the point clouds with a 5cm voxel size, then extract different feature descriptors to build the initial correspondence set as input. The inlier threshold $\tau$ is set to 10cm.

\noindent\textbf{Evaluation metrics}. Following DGR~\cite{choy2020deep}, {we use} three evaluation metrics,  
namely (1) \textit{Registration Recall (RR)}, 
the percentage of successful alignment whose rotation error and translation error are below some thresholds,  (2) \textit {Rotation Error (RE)}, and (3) \textit{Translation Error (TE)}. {\textit{RE} and \textit{TE} are}
defined as 
\begin{equation}
\setlength\abovedisplayskip{0.1cm}
	\text{RE}(\mathbf{\hat R}) = \arccos \frac{\text{Tr}(\mathbf{\hat R^T}\mathbf{R^*})-1}{2},~~~
	\text{TE}(\mathbf{\hat t}) = \left\Vert \mathbf{\hat t - t^*}\right\Vert_2,
    \label{eq:te}
\end{equation}
where $\mathbf{R^*}$ and $\mathbf{t^*}$ denote the ground-truth rotation and translation, respectively, and the average \textit{RE} and \textit{TE} are computed only on successfully registered pairs. 
Besides, we also report the intermediate outlier rejection results, including \textit{Inlier Precision (IP)}=$\frac{\#kept~inliers}{\#kept~matches}$ and \textit{Inlier Recall (IR)}=$\frac{\#kept~inliers}{\#inliers}$, which are {particularly} introduced to 
evaluate the outlier rejection module. For \textit{RR}, one registration result is considered successful if the \textit{TE} is less than 30cm and the \textit{RE} is less than 15\textdegree. For a fair comparison, we report two sets of results by combining different outlier rejection algorithms with the learned descriptor FCGF~\cite{Choy_2019_ICCV} and hand-crafted descriptor FPFH~\cite{rusu2009fast}, respectively.



\begin{table*}[th]
    \centering
    \vspace{-0.6cm}
    \setlength{\abovecaptionskip}{0.10cm}
 	\setlength{\belowcaptionskip}{-0.20cm}
    \resizebox{0.98\textwidth}{!}{
        \begin{tabular}{l|ccccccc|ccccccc}
        \Xhline{2\arrayrulewidth}
         & \multicolumn{7}{c|}{\textbf{FCGF}~(learned descriptor)} & \multicolumn{7}{c}{\textbf{FPFH}~(traditional descriptor)} \\
         & {RR(\%$\uparrow$)} & {RE(\textdegree$\downarrow$)} & {TE(cm$\downarrow$)} & {IP(\%$\uparrow$)} & {IR(\%$\uparrow$)} & {F1(\%$\uparrow$)} & {Time(s)} & {RR(\%$\uparrow$)} & {RE(\textdegree$\downarrow$)} & {TE(cm$\downarrow$)} & {IP(\%$\uparrow$)} & {IR(\%$\uparrow$)} & {F1(\%$\uparrow$)} & {Time(s)} \\
        \hline
         \textbf{FGR \cite{zhou2016fast}}         & 78.56 & 2.82 & 8.36 & - & - & - & 0.76 & 40.67 & 3.99 & 9.83 & - & - & - & 0.28 \\
         \hdashline
         \textbf{SM \cite{leordeanu2005spectral}}          & 86.57 & 2.29 & 7.07 & {81.44} & 38.36 & 48.21 & 0.03 & 55.88 & 2.94 &  8.15 & 47.96 & 70.69 & 50.70 & 0.03 \\
         \textbf{TEASER \cite{Yang20arXivTEASER}}      & 85.77 & 2.73 & 8.66 & \textbf{82.43} & 68.08 & 73.96 & 0.11 & 75.48 & 2.48 & 7.31 & \textbf{73.01} & 62.63 & 66.93 & 0.03 \\
         \textbf{GC-RANSAC-100k \cite{barath2018graph}} & 92.05 & 2.33 & 7.11 & 64.46 & \textbf{93.39} & 75.69 & 0.47 & 67.65 & 2.33 & 6.87 & 48.55 & 69.38 & 56.78 & 0.62 \\
         \textbf{RANSAC-1k \cite{fischler1981random}}   & 86.57 & 3.16 & 9.67 & 76.86 & 77.45 & 76.62 & 0.08 & 40.05 & 5.16 & 13.65 & 51.52 & 34.31 & 39.23 & 0.08 \\
         \textbf{RANSAC-10k}  & 90.70 & 2.69 & 8.25 & 78.54 & 83.72 & 80.76 & 0.58 & 60.63 & 4.35 & 11.79 & 62.43 & 54.12 & 57.07 & 0.55 \\
         \textbf{RANSAC-100k}   & 91.50 & 2.49 & 7.54 & 78.38 & 85.30 & 81.43 & 5.50 & 73.57 & 3.55 & 10.04 & 68.18 & 67.40 & 67.47 & 5.24 \\
         \textbf{RANSAC-100k refine}  & 92.30 & 2.17 & 6.76 & 78.38 & 85.30 & 81.43 & 5.51 & 77.20 & 2.62 & 7.42 & 68.18 & 67.40 & 67.47 & 5.25 \\
         \hdashline
         \textbf{3DRegNet \cite{pais20203dregnet}}  & 77.76 & 2.74 & 8.13 & 67.34 & 56.28 & 58.33 & 0.05 & 26.31 & 3.75 & 9.60 & 28.21 & 8.90 & 11.63 & 0.05 \\
         \textbf{DGR w/o s.g. \cite{choy2020deep}} & 86.50 & 2.33 & 7.36 & 67.47 & 78.94 & 72.76 & 0.56 & 27.04 & 2.61 & 7.76 & 28.80 & 12.42 & 17.35 & 0.56  \\
         \textbf{DGR \cite{choy2020deep}}        & 91.30 & 2.40 & 7.48 & 67.47 & 78.94 & 72.76 & 1.36 & 69.13 & 3.78 & 10.80 & 28.80 & 12.42 & 17.35 & 2.49  \\
         \textbf{\Name}      & \textbf{93.28} & \textbf{2.06} & \textbf{6.55} & 79.10 & {86.54} & \textbf{82.35} & 0.09 & \textbf{78.50} & \textbf{2.07} & \textbf{6.57} & 68.57 & \textbf{71.61} & \textbf{69.85} & 0.09  \\
        \Xhline{2\arrayrulewidth}
 		\end{tabular}
       }
    \caption{Registration results on 3DMatch. 
 \textit{RANSAC-100k refine} represents RANSAC with 100k iterations, followed by the proposed post{-}refinement step. \textit{DGR w/o s.g.} represents DGR~\cite{choy2020deep} without the safeguard mechanism~(RANSAC). {The \textit{Time} columns report the average time cost during testing, excluding} 
 the construction of initial {input} correspondences.
    }
    
    \label{tab:3dmatch}
\end{table*}

\noindent\textbf{Baseline methods.} 
We first select four representative traditional methods: FGR~\cite{zhou2016fast}, SM~\cite{leordeanu2005spectral}, RANSAC~\cite{fischler1981random}, and GC-RANSAC~\cite{barath2018graph}, as well as the state-of-the-art geometry-based method TEASER~\cite{Yang20arXivTEASER}. For learning-based methods, we choose 3DRegNet~\cite{pais20203dregnet} and DGR~\cite{choy2020deep} as the baselines, since they also focus on the outlier rejection step {for point cloud} registration. We also report the results of DGR without RANSAC~({i.e., without the so-called safeguard mechanism})
to better compare the weighted least-square{s} solutions. We carefully tune each method to achieve the best results on the evaluation dataset for a fair comparison. More details can be found in the supplementary~\ref{supp:baseline}.

\noindent\textbf{Comparisons with the state-of-the-arts.} We compare our \Name~with the baseline methods on 3DMatch. As shown in Table~\ref{tab:3dmatch}, all the evaluation metrics are reported in two settings: input putative correspondences constructed by  FCGF~(left columns) and FPFH~(right columns). \Name~achieves the best \textit{Registration Recall} as well as the lowest average \textit{TE} and \textit{RE} in both settings. {More statistics can be found in the supplementary~\ref{supp:auc}.}

\noindent\textbf{Combination with FCGF descriptor.} Compared with the {learning-based baselines}, \Name~surpasses the second best method, i.e., DGR, by more than \textbf{9\%} 
in terms of \textit{F1 score}, indicating the effectiveness of our outlier rejection algorithm. 
Besides, although DGR {is only slightly worse than \Name}~in  
\textit{Registration Recall}, it is noteworthy that more than 35\% (608/1623) registration pairs are marked as failure and solved by RANSAC~(safeguard mechanism). If no safeguard mechanism is applied, DGR only achieves 86.5\% \textit{Registration Recall}.

{Different from the conclusion in \cite{choy2020deep}, our experiments indicate that}
RANSAC still shows competitive results when combined with a powerful descriptor FCGF. Nevertheless, our method is about \textbf{60 times} faster than RANSAC-100k while achieving even higher \textit{Registration Recall}. We also report the performance of RANSAC with the proposed post-refinement step
 to clearly demonstrate the superiority of our outlier rejection module. 
{SM and TEASER achieve slightly better \textit{Inlier Precision} than \Name, however,
they} 
have much lower \textit{Inlier Recall}
(38.36\% and 68.08\% vs. 86.54\% (Ours)). We thus conclude that \Name~achieves
a better trade-off between precision and recall.

\noindent\textbf{Combination with {FPFH} descriptor.}
We further evaluate all the outlier rejection methods equipped with the traditional descriptor, FPFH. Note that for testing learnable outlier rejection methods {including \Name},  
 we directly re-use the model trained with the FCGF descriptor without fine-tuning, 
{since it is expected that the outlier rejection networks are seamlessly compatible with different feature descriptors.} As shown in Table \ref{tab:3dmatch}, the superiority of \Name~becomes more obvious when evaluated with the FPFH, where \Name~achieves 78.5\% in \textit{Registration Recall} and remarkably surpasses the 
competitors.
RANSAC-1k and RANSAC-10k perform significantly worse since the outlier ratios are  
much higher when using FPFH to build the input correspondences. RANSAC-100k with the post-refinement step still achieves the second best performance at the cost of the high computation time. 
{In summary,} all the other methods suffer from larger performance degradation than \Name~when equipped with a weaker descriptor, 
strongly demonstrating the robustness of \Name~to the input correspondence{s} generated by different feature descriptors. 


%% file: Experiment_part2.tex
\subsection{Generalization to Outdoor Scenes}
\label{subsec:kitti}

\begin{table}[t]
\scriptsize
    \centering
    \vspace{-0.25cm}
 	\setlength{\abovecaptionskip}{0.10cm}
 	\setlength{\belowcaptionskip}{-0.20cm}
    \resizebox{0.45\textwidth}{!}{
        \begin{tabular}{l|cccccc}
        \Xhline{1.5\arrayrulewidth}
         & {RR($\uparrow$)} & {RE($\downarrow$)} & {TE($\downarrow$)} & {F1($\uparrow$)} & {Time} \\
        \hline
         \textbf{SM \cite{leordeanu2005spectral}} & 79.64 & 0.47 & 12.15 & 56.37 & 0.18 \\
         \textbf{RANSAC-1k \cite{fischler1981random}}    & 11.89 & 2.51 & 38.23 & 14.13 & 0.20 \\
         \textbf{RANSAC-10k}   & 48.65 & 1.90 & 37.17 & 42.35 & 1.23 \\
         \textbf{RANSAC-100k}  & 89.37 & 1.22 & 25.88 & 73.13 & 13.7 \\
        \textbf{DGR \cite{choy2020deep}} & 73.69 & 1.67 & 34.74 & 4.51 & 0.86\\
         \textbf{\Name} & 90.27 & \textbf{0.35} & \textbf{7.83} & 70.89 & 0.31  & \\ 
         \textbf{DGR~re-trained}          & 77.12 & 1.64 & 33.10 & 27.96 & 0.86 \\
         \textbf{\Name~re-trained} & \textbf{98.20} & \textbf{0.35} & {8.13} & \textbf{85.54} & 0.31  \\

        \Xhline{1.5\arrayrulewidth}
        \end{tabular}
    }
   
    \caption{Registration results on KITTI under FPFH setting.}
    \label{tab:kitti_fpfh}
\end{table}

In order to evaluate the generalization of \Name~to new datasets and unseen domains, we evaluate on a LiDAR outdoor dataset, namely the KITTI odometry dataset, using the model trained on 3DMatch. We follow the same data splitting strategy in~\cite{Choy_2019_ICCV, choy2020deep} for a fair comparison.
We use 30cm voxel size 
and set the inlier threshold $\tau$ to 60cm. The evaluation metrics are the same as {those used in} the indoor setting with a 60cm \textit{TE} threshold and a 5\textdegree~\textit{RE} threshold.

\noindent\textbf{Comparisons with the state-of-the-arts.}  We choose SM, DGR, and RANSAC as the baseline methods, and combine them with the FPFH descriptor. We choose FPFH because the results with FCGF are more or less saturated.~(The results with FCGF can be found in the supplementary~\ref{supp:exp}.)
We report two sets of results for DGR and \Name~obtained when trained from scratch (labelled
 ``re-trained") and pre-trained on 3DMatch (no extra label). As shown in Table~\ref{tab:kitti_fpfh}, \Name~trained on 3DMatch still gives competitive results, demonstrating {its} strong generalization ability on the unseen dataset. When re-trained from scratch, \Name~can be further improved and outperform the baseline approaches by a significant margin. 

\begin{table}[tb]
 \scriptsize
 	\vspace{-0.25cm}
 	\setlength{\abovecaptionskip}{0.10cm}
 	\setlength{\belowcaptionskip}{-0.20cm}
    \centering
    \resizebox{0.45\textwidth}{!}{
        \begin{tabular}{l|cccccc}
        \Xhline{1.5\arrayrulewidth}
         & {Living1} & {Living2} & {Office1} & {Office2} & {AVG}\\
        \hline

         \textbf{ElasticFusion \cite{whelan2015elasticfusion}}    & 66.61 & 24.33 & \textbf{13.04} & 35.02 & 34.75 \\
         \textbf{InfiniTAM \cite{kahler2016real}}        & 46.07 & 73.64 & 113.8 & 105.2 & 84.68 \\
         \textbf{BAD-SLAM\cite{schops2019bad}}         & fail  & 40.41 & 18.53 & 26.34 &  -\\
         \textbf{Multiway + FGR \cite{zhou2016fast}}              & 78.97 & 24.91 & 14.96 & 21.05 & 34.98 \\
         \textbf{Multiway + RANSAC \cite{fischler1981random}}           & 110.9 & 19.33 & 14.42 & 17.31 & 40.49 \\
         \textbf{Multiway + DGR \cite{choy2020deep}}              & 21.06 & 21.88 & 15.76 & 11.56 & 17.57 \\
		  \textbf{Multiway + \Name}            & \textbf{20.25} & \textbf{15.58} & {13.56} & \textbf{11.30} & \textbf{15.18}  \\

        \Xhline{1.5\arrayrulewidth}
        \end{tabular}
    }
   
    \caption{ATE(cm) on Augmented ICL-NUIM. {The last column is the average ATE over all scenes.} {Since BAD-SLAM fails on one scene, we do not report its average ATE.}
    }
    \label{tab:multi}
\end{table}

\subsection{Multiway Registration}
\label{subsec:multi}
For evaluating multiway registration, we use Augmented ICL-NUIM dataset~\cite{choi2015robust}, which augments each synthetic scene~\cite{handa2014benchmark} with 
a realistic noise model.
To test the generalization ability,  we again use the models trained on 3DMatch without fine-tuning. Following \cite{choy2020deep}, we first perform pairwise registration using \Name~with FPFH descriptor {to obtain the initial poses}, then optimize the poses
using pose graph optimization~\cite{kummerle2011g} implemented in Open3D~\cite{zhou2018open3d}. We report the results of baseline methods presented in \cite{choy2020deep}. The \textit{Absolute Trajectory Error~(ATE)} is reported as the evaluation metric.
As shown in Table~\ref{tab:multi}, our method achieves the {lowest} 
average \textit{ATE} over {three of the four tested scene types}. 

\subsection{Ablation Studies}
\label{subsec:ablation}

\noindent\textbf{Ablation on {feature encoder}. 
} To study the effectiveness of the proposed \nonlocal~module, we conduct extensive ablation experiments on 3DMatch. Specifically, we compare (1) \textbf{PointCN}~(3D version of~\cite{moo2018learning}, {which is the feature extraction module adopted by {3DRegNet}~\cite{pais20203dregnet}});
(2) \textbf{Nonlocal}~(the \nonlocal~module without {the} spatial term, i.e., the same operator as in~\cite{wang2018non}); 
and (3) \textbf{\nonlocal}~(the proposed operator). All the above methods are combined either with a classification layer~\cite{choy2020deep, pais20203dregnet} or a proposed \nsm~layer, resulting {in} six combinations in total. Other training or testing settings remain unchanged for a fair comparison. 

\begin{figure}[t]
	\centering
\vspace{-0.35cm}
 	\setlength{\abovecaptionskip}{0.10cm}
 	\setlength{\belowcaptionskip}{-0.20cm}
	\hspace{-0.5cm}
	\includegraphics[height=4.0cm]{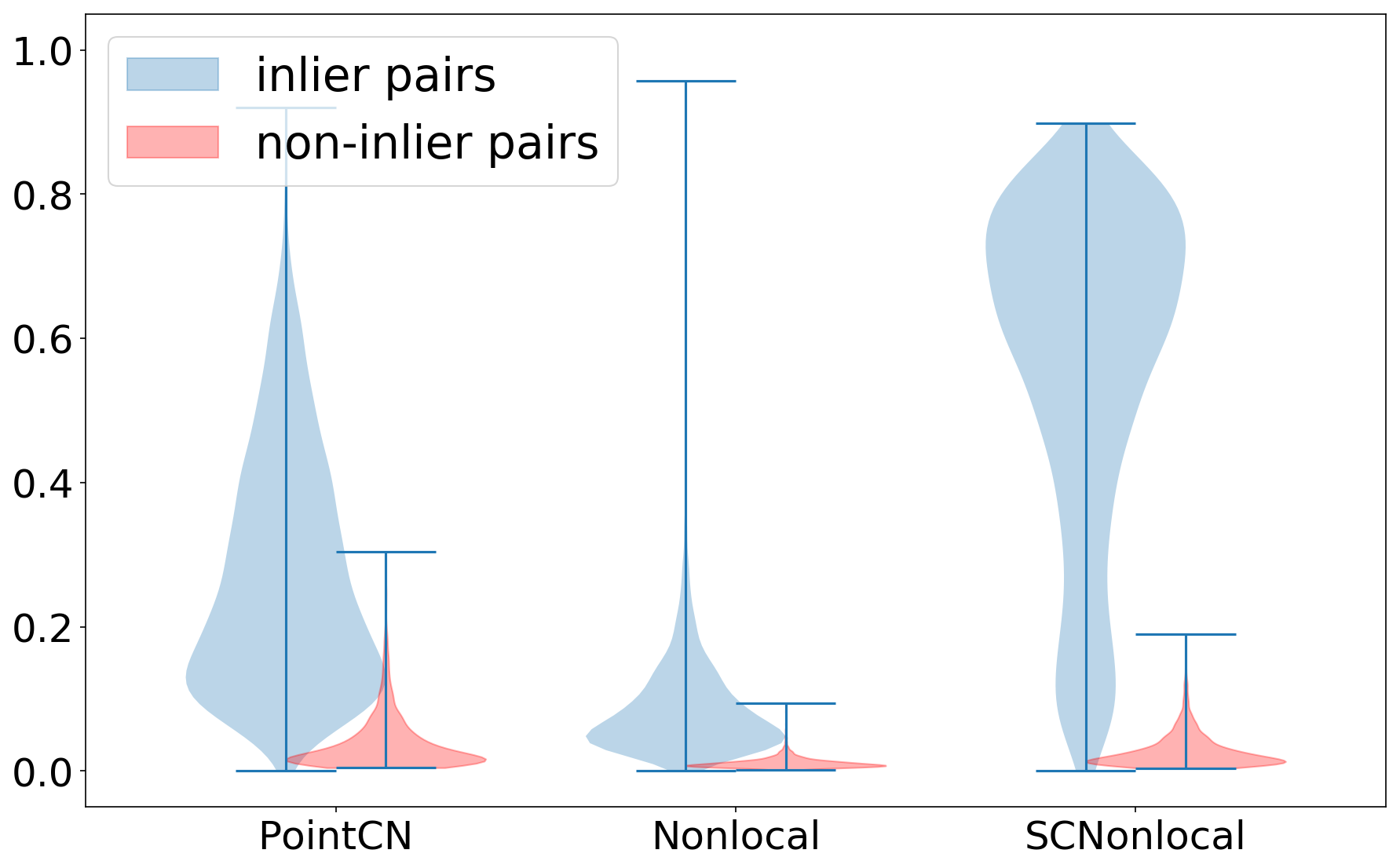}
    \caption{The distribution of feature similarity of inlier pairs and non-inlier pairs~(i.e. at least one outlier in the pair).}
    \label{fig:abl_nonlocal}
\end{figure}

{As shown in Table~\ref{tab:ablation_feat}, the proposed \nonlocal~module consistently improves the registration results across all the settings and metrics.} \xy{The spatial term plays a critical role in the \nonlocal~module, without which the Nonlocal module performs drastically worse.}
\xy{Furthermore, we compute the feature similarity defined in Eq.~\ref{eq:feat_sm} between each pair of correspondences and {plot} the distribution in Fig.~\ref{fig:abl_nonlocal}. With the~\nonlocal~module, the similarity of the inlier pairs is concentrated near 0.8 and is generally much larger than that of the non-inlier pairs. This implies that inliers are closer to each other in the embedding space. In contrast, for the baseline methods, the inliers are less concentrated, i.e., the average similarity between inliers is low.}

\begin{table}[t]
 \scriptsize
    \centering
   	\setlength{\abovecaptionskip}{0.10cm}
	\setlength{\belowcaptionskip}{-0.2cm}
    \resizebox{0.450\textwidth}{!}{
        \begin{tabular}{lc|cccccc}
        \Xhline{1.5\arrayrulewidth}
         & & {RR($\uparrow$)} & {IP($\uparrow$)} & {IR($\uparrow$)} & {F1($\uparrow$)} & Time  \\
        \hline
        \textbf{PointCN}   & \textbf{+ classifier}  & 78.19 & 58.05 & 39.59 & 42.65 & 0.04 \\
        \textbf{Nonlocal}  & \textbf{+ classifier}     & 83.30 & 65.49 & 67.13 & 64.28 & 0.07 \\
        \textbf{\nonlocal} & \textbf{+ classifier}    & 88.17 & 74.74 & 77.86 & 75.04 & 0.07 \\
         
         \hdashline
         \textbf{PointCN}  &\textbf{+ \nsm}  & 92.48 & 78.48 & 82.10 & 79.98 & 0.06 \\
         \textbf{Nonlocal} &\textbf{+ \nsm}     & 92.54 & 78.68 & 83.13 & 80.58 & 0.09 \\
         \textbf{\nonlocal} &\textbf{+ \nsm}   & \textbf{93.28} & \textbf{79.10} & \textbf{86.54} & \textbf{82.35} & 0.09 \\
        \Xhline{1.5\arrayrulewidth}
        \end{tabular}
    }
    
    \caption{Ablation experiments of \nonlocal~module. Row{s} 1-3 and Row{s} 4-6 show the registration results of different feature extractors combined with a classification layer and the neural spectral matching module, respectively.
    }
    \label{tab:ablation_feat}
\end{table}

\noindent\textbf{Ablation on spectral matching.} 
We further conduct ablation experiments to demonstrate the importance of \nsm~module.
{As shown in Table~\ref{tab:ablation_sm},
the comparison between Rows 1 and 2 shows that augmenting the traditional SM with neural feature consistency
 notably improves the result. 
For \textbf{+seeding}, we adopt the neural spectral matching over multiple correspondence subsets found by the {feature-space} kNN search from highly confident \textit{seeds}, and determine the best transformation that maximizes the geometric consensus. This significantly boosts the performance because it is easier to find the inlier clusters for the consistent correspondence subsets.

\begin{table}[t]
 \scriptsize
    \centering
    \vspace{-0.35cm}
    \setlength{\abovecaptionskip}{0.10cm}
    \setlength{\belowcaptionskip}{-0.20cm}
    \resizebox{0.40\textwidth}{!}{
        \begin{tabular}{l|ccccc}
        \Xhline{1.5\arrayrulewidth}
    
         & {RR($\uparrow$)} & {RE($\downarrow$)} & {TE($\downarrow$)} & {F1($\uparrow$)} & {Time}  \\
        \hline
		 \textbf{Traditional SM}         & 86.57 & 2.29 & 7.07 & 48.21 & 0.03 \\
         \textbf{ + neural}              & 88.43 & 2.21 & 6.91 & 48.88 & 0.06 \\
         \textbf{ + seeding}            & 92.91 & 2.15 & 6.72 & 82.35 & 0.08 \\
         \textbf{ + refine}     & \textbf{93.28} & \textbf{2.06} & \textbf{6.55} & \textbf{82.35} & 0.09 \\
         \hdashline
         \textbf{ w/o $L_{sm}$}  & 92.61 & 2.07 & 6.75 & 81.58 & 0.09 \\

        \Xhline{1.5\arrayrulewidth}
        \end{tabular}
    }
    
    \caption{Ablation experiments of \nsm~module. 
    {Note that every row {with `+'} represents the previous row equipped with the new component.} 
	\textbf{+refine} is our full model.
    {The} last row is the full model trained without $L_{sm}$.} 
    \label{tab:ablation_sm}
\end{table}

\subsection{Qualitative Results}
As shown in Fig.~\ref{fig:vis_best}, \Name~{is} 
robust to extremely high outlier ratios. 
Please refer to the supplementary~\ref{supp:vis}} for more qualitative results.

\begin{figure}[tb]
	\setlength{\abovecaptionskip}{0.10cm}
	\setlength{\belowcaptionskip}{-0.35cm}
    \includegraphics[width=8.5cm]{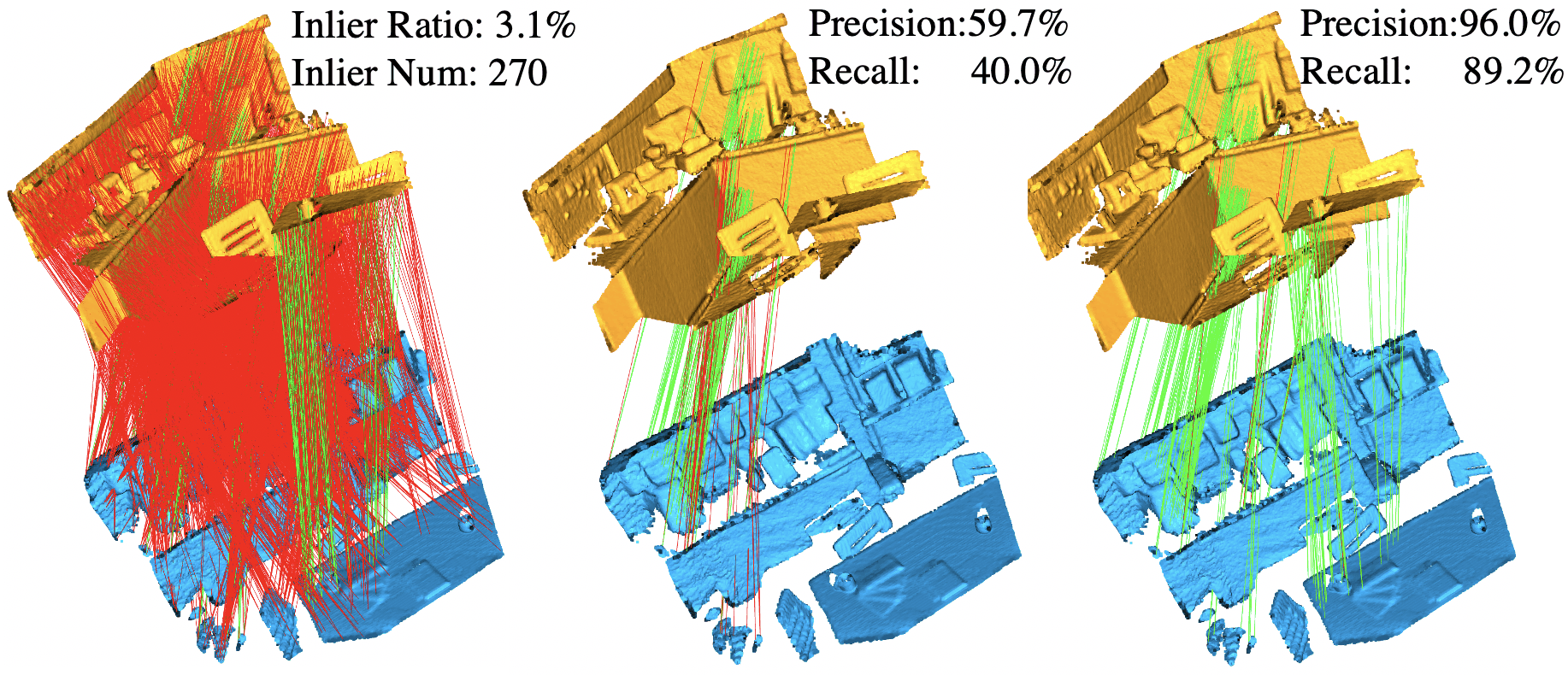} 
    \hspace{2cm}
   	\includegraphics[width=8.5cm]{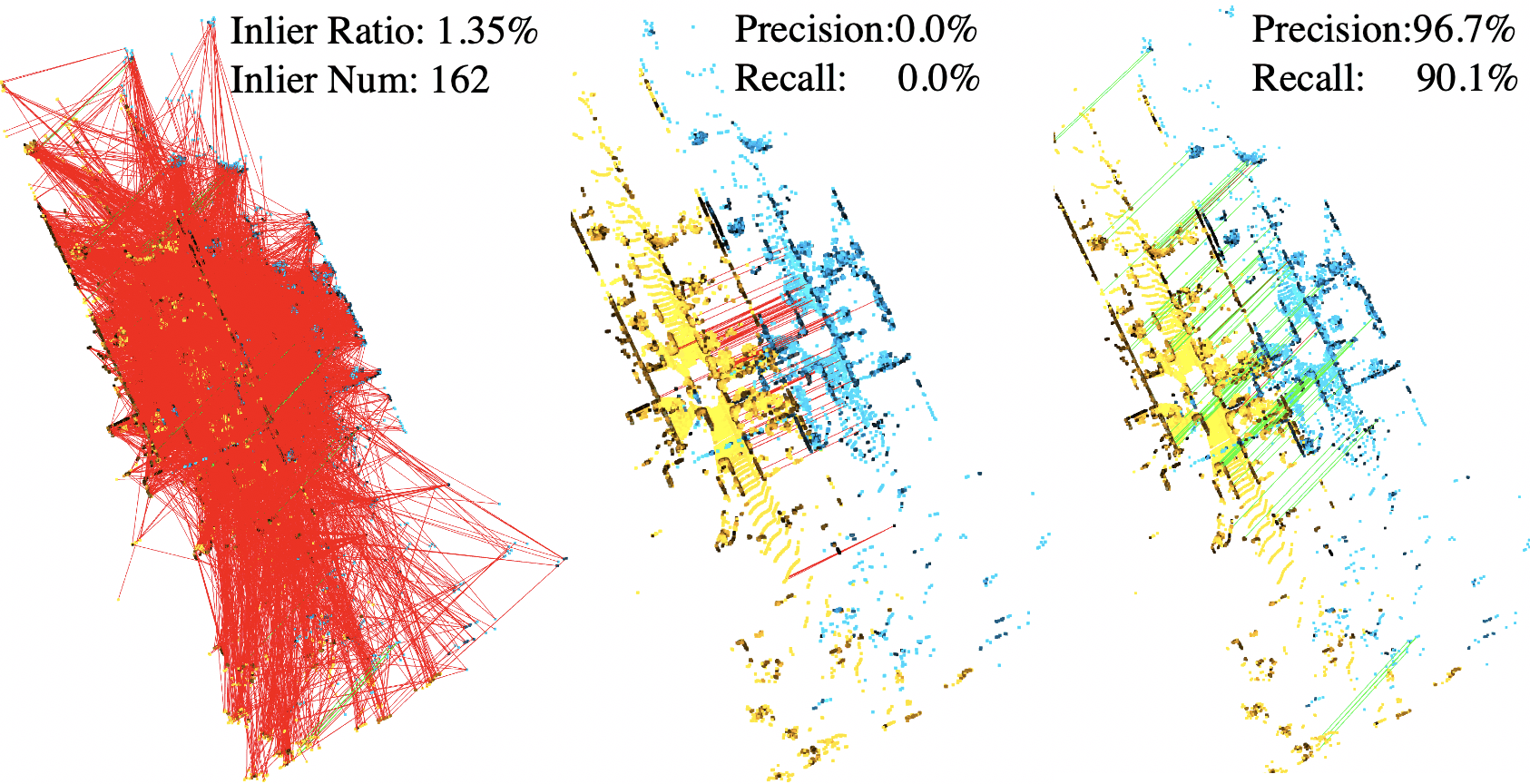}
     \caption{
     Visualization of outlier rejection results on {examples with high outlier ratios from 3DMatch~(first row) and KITTI~(second row)}. From left to right: input correspondences, results of RANSAC-100k, and results of \Name. 
     }
    \label{fig:vis_best}
\end{figure}

%% file: Conclusion.tex
We have designed a novel 3D outlier rejection network that explicitly incorporates \textit{spatial consistency} established by Euclidean transformations.
{We have proposed} a spatial-consistency guided nonlocal module~(\nonlocal) and a neural spectral matching module~(\nsm) for feature embedding and outlier pruning, respectively. {We further proposed a seeding mechanism to adopt the NSM module multiple times to boost the robustness under high outlier ratios.}
The extensive experiments on {diverse} datasets {showed that our method brings} 
remarkable improvement over the state-of-the-arts.
 Our method can also generalize to unseen domains and cooperate with different local descriptors seamlessly.  \\

\noindent\textbf{Acknowledgements.} This work is supported by Hong Kong RGC GRF 16206819, 16203518 and Centre for Applied Computing and Interactive Media (ACIM) of School of Creative Media, City University of Hong Kong.

%% file: Supplementary.tex
\subsection{Implementation Details of \Name}
\label{supp:post}

\noindent We provide additional information about the implementation and training details of our \Name.
The source code will be made publicly available after the paper gets accepted. 

\noindent\textbf{Post-refinement.}
Alg.~\ref{alg:post_refine} shows the pseudo-code of our post-refinement step. Inspired by \cite{bergstrom2014robust}, we iteratively alternate between {weighing} 
the correspondences and computing the transformation, {to improve the accuracy of the transformation matrices.}
The inlier threshold $\tau$ is set to 10cm and 60cm for 3DMatch and KITTI, respectively. We set the maximum iteration number to 20.

\begin{algorithm}[htb]
\SetKwInput{KwInput}{Input}                
\SetKwInput{KwOutput}{Output} 
\SetKwInput{KwParameter}{Parameter} 
\DontPrintSemicolon
  
  \KwInput{$\mathbf{\hat R, \hat t}$: initial transformation; $\mathbf{X, Y}$}
  \KwOutput{$\mathbf{\hat R, \hat t}$: refined transformation.}
  \KwParameter{$\tau$.}
  \If{$iter < max_{iter}$}
    {	
    	\color{red}{\# Compute the residual and the inlier num.}
		\color{black}$res_i = \vert\vert \mathbf{\hat R}\bm{x_i} + \mathbf{\hat t} - \bm{y_i} \vert\vert_2 $ \\
        \color{black}$w_i = \llbracket res_i < \tau \big \rrbracket$ \\
    	\color{black}$num = \sum w_i$    
    
        \color{red}{\# If inlier num does not change, then stop.} \\
        \color{black}\If{$\Delta num = 0$}    {break}
		 \Else{
		
		\color{red}{\# Compute the weighting term.} \\
		\color{black}$\phi_i = (1 + (\frac{res_i}{\tau})^2)^{-1}$ \\
        \color{red}{\# Estimate transformation.}  \\
        \color{black}$\mathbf{\hat R, \hat t} = \arg\min_{\mathbf{R, t}} \sum_{i}^{N} \phi_iw_i\left\Vert \mathbf{R}\bm{x_i} + \mathbf{t} - \bm{y_i}  \right\Vert^2$ \\
        }
        
        \color{black}$iter = iter + 1$
    }
    \Else
    {
    	break
    }
    
\caption{Post-Refinement Algorithm}
\label{alg:post_refine}
\end{algorithm}

\noindent\textbf{Calculation of M.} In Sec.~\ref{subsec:nsm} of the main text, 
we calculate the compatibility between correspondences by multiplying the spatial consistency term and the feature similarity term mainly because of its simplicity and good performance. Other fusion schemes such as the weighted arithmetic average and weighted geometric average can also be used to
{define the compatibility metric}. 
We have explored several alternatives {but} 
{found only a} 
marginal performance difference.

\noindent\textbf{Power iteration algorithm.}
The power iteration algorithm can compute the leading eigenvector $\bm{e}$ of the matrix  $\mathbf{M}$ in several iterations. For $\mathbf{M} \in \mathbb{R}^{k\times k}$, the power iteration operator is
\begin{equation}
\setlength\abovedisplayskip{5pt}\setlength\belowdisplayskip{5pt}
    \bm{e}^{\text{iter+1}} = \frac{\mathbf{M}\bm{e}^{\text{iter}}}{\left\Vert \mathbf{M}\bm{e}^{\text{iter}} \right\Vert}.
    \label{eq:power}
\end{equation}
We initialize $\bm{e^0} = \textbf{1}$. By iterating Eq.~\ref{eq:power} until convergence, we get the vector $\bm{e}$, whose elements can take real values in $[0,1]$. In practice, we find that the power iteration algorithm usually converges in fewer than five iterations.

\noindent\textbf{Data augmentation.} During training, we apply data augmentation, including adding Gaussian noise with standard deviation of 0.005, random rotation angle $\in [0^{\circ}, 360^{\circ})$ around an arbitrary axis, and random translation $\in [-0.5m, 0.5m]$ around each axis.

\noindent\textbf{Hyper-parameters.} {The hyper-parameter $\sigma_d$ controls the sensitivity to length difference, serving as a pairwise counterpart of the unary inlier threshold $\tau$. The larger $\sigma_d$, the more length difference between two pairs of correspondences we can accommodate. It is set manually {for a specific scene} and kept fixed. {Picking a scene-specific value of $\sigma_d$ is easy due to its clear physical meaning. } 
However, $\sigma_f$ controlling the sensitivity to feature difference has no clear physical meaning. 
{We thus}
leave $\sigma_f$ to be learned by the network.
}

\subsection{Implementation Detail of Baseline Methods}
\label{supp:baseline}
The baseline methods RANSAC~\cite{fischler1981random} and FGR~\cite{zhou2016fast} have been 
implemented in Open3D~\cite{zhou2018open3d}. For GC-RANSAC~\cite{barath2018graph} and TEASER~\cite{Yang20arXivTEASER}, we use the official implementations. Note that we use TEASER with reciprocal check; otherwise, it takes an extremely long time for testing when the number of input correspondences becomes large. For DGR~\cite{choy2020deep}, we use {its} 
official implementation and the released pre-trained model.
{Due to the unsatisfactory results of publicly released code,}
we re-implement SM~\cite{leordeanu2005spectral} and 3DRegNet~\cite{pais20203dregnet}, with the implementation details as follows. 


\noindent\textbf{Spectral {m}atching.} Traditional spectral matching~\cite{leordeanu2005spectral} uses a greedy algorithm based on {a} 
one-to-one mapping constraint to discretize the leading eigenvector into the inlier/outlier labels. However, the greedy algorithm {often} does not show satisfactory performance in real cases. For example, if the input correspondences are pre-filtered by reciprocal check, the greedy algorithm could not reject any correspondences since all of them already satisfy the one-to-one mapping constraint. {The} Hungarian algorithm~\cite{munkres1957algorithms} can also be used for discretization but provides results similar to the greedy algorithm.
 In our work, we simply select 10\% {of the input correspondences} with the highest confidence values as the inlier set. {This approach} 
 empirically shows to be effective throughout our experiments. Then the transformation between two point clouds can be estimated using the selected correspondences.

\noindent\textbf{3DRegNet.} We keep the network architecture proposed in 3DRegNet~\cite{pais20203dregnet} and train it on 3DMatch using the same settings as \Name. However, as observed in~\cite{choy2020deep}, {3DRegNet} does not converge during training and the registration block cannot 
produce reasonable results. We speculate that directly regressing the pose results in the poor performance due to the non-linearity of the rotation space~\cite{peng2019pvnet, he2020pvn3d}. Thus we regard the output of the classification block as the inlier confidence and 
use the confidence as the weight for weighted least-squares fitting. We then train the network using the classification loss only, since we find the registration loss does not improve the performance. The modified 3DRegNet becomes a 3D variant of PointCN~\cite{moo2018learning} and achieves reasonable results.

\subsection{Time Complexity Analysis}
We report the average runtime of each component in the proposed pipeline on the 3DMatch test set (roughly 5k putative correspondence per fragment) in Table~\ref{tab:runtime}. The reported times are measured using an Intel Xeon 8-core~2.1GHz~CPU~(E5-2620) and an NVIDIA GTX1080Ti GPU. 

\begin{table}[h]
    \centering
    \setlength{\abovecaptionskip}{0.10cm}
    \setlength{\belowcaptionskip}{-0.20cm}
    \resizebox{0.45\textwidth}{!}{
        \begin{tabular}{ccccc}
        \Xhline{1.5\arrayrulewidth}
         {\nonlocal} & {Seed Selection} & {\nsm} & {Post Refine} & {Overall}\\
        \hline
          62.0 & 2.0 & 14.4 & 11.1 & 89.5\\
        \Xhline{1.5\arrayrulewidth}
        \end{tabular}
    }
   
    \caption{Runtime of each component in \textbf{milli-seconds}, averaged over 1,623 test pairs of 3DMatch. The time of hypothesis selection is included in the NSM module.}
    \label{tab:runtime}
\end{table}

\subsection{Additional Statistics}
\label{supp:auc}
We report the area under cumulative error curve~(AUC) of the rotation and translation errors defined in Eq.~\ref{eq:te} at different thresholds, as shown in Table~\ref{tab:auc}. \Name~consistently outperforms the state-of-the-arts 
on both the AUC of the \textit{Rotation Error} (RE) and \textit{Translation Error} (TE). 
\begin{table}[th]
    \centering
        \setlength{\abovecaptionskip}{0.10cm}
    \setlength{\belowcaptionskip}{-0.20cm}
    \resizebox{0.48\textwidth}{!}{
        \begin{tabular}{l|ccc|cccccc}
        \Xhline{1.5\arrayrulewidth}
         & \multicolumn{3}{c|}{\textbf{RE AUC}} & \multicolumn{6}{c}{\textbf{TE AUC}} \\
         & {$5^{\circ}$} & {$10^{\circ}$} & {$15^{\circ}$} & {$5$cm} & {$10$cm} & {$15$cm} & {$20$cm} & {$25$cm} & {$30$cm} \\
        \hline
         \textbf{SM} & 50.14 & 67.24 & 74.37 & 16.29 & 35.98 & 48.61 & 56.90 & 62.57 & 66.67\\
         \textbf{DGR} & 50.22 & 69.98 & 77.78 & 14.13 & 35.28 & 49.32 & 58.50 & 64.74 & 69.19\\
         \textbf{RANSAC} & 49.99 & 70.43 & 78.31 & 12.16 & 33.15 & 47.99 & 57.81 & 64.33 & 68.95\\
         \textbf{GC-RANSAC} & 52.81 & 71.56 & 78.90 & 15.33 & 36.77 & 50.94 & 59.95 & 65.94 & 70.19\\
         \textbf{\Name}  & \textbf{57.32} & \textbf{74.85} & \textbf{81.50} & \textbf{17.85} & \textbf{40.63} & \textbf{54.56} & \textbf{63.32} & \textbf{69.02} & \textbf{73.00} \\

        \Xhline{1.5\arrayrulewidth}
 		\end{tabular}
       }
    \caption{Registration results on 3DMatch. We calculate the exact AUC following \cite{sarlin2020superglue}:   {the higher, the better}. We run 100k iterations for both RANSAC and GC-RANSAC.}

    \label{tab:auc}
\end{table}

We also report the scene-wise registration results of our method on 3DMatch in Table~\ref{tab:scenewise}. 
\begin{table}[h]
    \centering
    \setlength{\abovecaptionskip}{0.10cm}
    \setlength{\belowcaptionskip}{-0.20cm}
    \resizebox{0.45\textwidth}{!}{
        \begin{tabular}{l|ccccccc}
        \Xhline{1.5\arrayrulewidth}
         & {RR(\%)} & {RE(\textdegree)} & {TE(cm)} & {IP(\%)} & {IR(\%)} & {F1(\%)}\\
        \hline
        
         \textbf{Kitchen} & 98.81 & 1.67 & 5.12 & 80.57 & 88.83 & 84.26 \\
		\textbf{Home1} & 97.44 & 1.87 & 6.45 & 83.34 & 88.91 & 85.88 \\
		\textbf{Home2} & 82.21 & 3.36 & 7.46 & 71.39 & 80.20 & 74.78 \\
		\textbf{Hotel1} & 98.67 & 1.88 & 6.04 & 83.96 & 91.48 & 87.38 \\
		\textbf{Hotel2} & 92.31 & 1.98 & 5.74 & 81.07 & 86.97 & 83.82 \\
		\textbf{Hotel3} & 92.59 & 2.00 & 5.87 & 82.65 & 88.57 & 85.03 \\
		\textbf{Study}  & 89.04 & 2.29 & 9.20 & 77.00 & 83.72 & 79.97 \\
		\textbf{Lab}    & 80.52 & 1.91 & 8.41 & 70.31 & 77.88 & 73.46 \\
        \Xhline{1.5\arrayrulewidth}
        \end{tabular}
    }
   
    \caption{Scene-wise statistics for \Name~on 3DMatch.}
    \label{tab:scenewise}
\end{table}

\subsection{Additional Experiments}
\label{supp:exp}

\noindent\textbf{Registration results on KITTI.} Due to the space limitation and the saturated performance under the FCGF setting, we only report the registration results on KITTI under the FPFH setting in the main text. Here we report the performance of all the methods combined with FCGF in Table~\ref{tab:kitti_fcgf}. For the learning-based models DGR and PointSM, we report the performance of the models trained from scratch~(labelled ``re-trained") and pre-trained on the indoor dataset 3DMatch~(no extra label) with the FCGF descriptor. 

\begin{table}[h]
    \centering
     \setlength{\abovecaptionskip}{0.10cm}
    \setlength{\belowcaptionskip}{-0.20cm}
    \resizebox{0.45\textwidth}{!}{
        \begin{tabular}{l|cccccc}
        \Xhline{1.5\arrayrulewidth}
         & {RR($\uparrow$)} & {RE($\downarrow$)} & {TE($\downarrow$)} & {F1($\uparrow$)} & {Time} \\
        \hline
%
         \textbf{SM} & 96.76 & 0.50 & \textbf{19.73} & 22.84 & 0.10 \\
         \textbf{RANSAC-1k}    & 97.12 &0.48 & 23.37 & 84.26 & 0.22 \\
         \textbf{RANSAC-10k}   & 98.02 & 0.41 & 22.94 & 85.05 & 1.43 \\
         \textbf{RANSAC-100k}  & \textbf{98.38} & 0.38 & 22.60 & \textbf{85.42} & 13.4 \\
         \textbf{DGR} & 95.14 & 0.43 & 23.28 & 73.60 & 0.86\\
         \textbf{\Name} & 97.84 & 0.33 & 20.99 & 85.29 & 0.31  & \\ 
        \textbf{DGR re-trained}          & 96.90 & 0.33 & 21.29 & 73.56 & 0.86 \\
         \textbf{\Name~re-trained} & 98.20 & \textbf{0.33} & {20.94} & 85.37 & 0.31  \\

        \Xhline{1.5\arrayrulewidth}
        \end{tabular}
    }
   
    \caption{Registration results on KITTI under the FCGF setting. The reported time numbers do not include the construction of initial correspondences.
    }
    \label{tab:kitti_fcgf}
\end{table}

\xy{
\noindent\textbf{Under low-overlapping cases.}
Recently, Huang et. al~\cite{huang2020predator} have constructed a low-overlapping dataset 3DLoMatch from the 3DMatch benchmark to evaluate the point cloud registration algorithms under low-overlapping scenarios. To demonstrate the robustness of our~\Name, we further evaluate our method on 3DLoMatch dataset and report the results\footnote{The computation of registration recall is slightly different with ours, we refer readers to \cite{huang2020predator} for more details.} in Table \ref{tab:3DLoMatch}. Note that we directly use the model trained on 3DMatch without fine-tuning and keep 5cm voxel for FCGF descriptor. All the other settings are the same with~\cite{huang2020predator} for a fair comparison.}

\begin{table}[h]
    \centering
    \setlength{\abovecaptionskip}{0.10cm}
    \setlength{\belowcaptionskip}{-0.20cm}
    \resizebox{0.48\textwidth}{!}{
        \begin{tabular}{l|cccccccc}
        \Xhline{1.5\arrayrulewidth}
           & {5000} & {2500} &  {1000}& {500} & {250} & {$\Delta$} \\
        \hline
        \textbf{FCGF\cite{Choy_2019_ICCV} + RANSAC}    & 35.7 & 34.9 & 33.4 & 31.3 & 24.4 &  -\\ 
		\textbf{FCGF\cite{Choy_2019_ICCV} + \Name}     & 52.0 & 51.0 & 45.2 & 37.7 & 27.5 & +10.74\\
        \textbf{Predator\cite{huang2020predator} + RANSAC}    & 54.2 & 55.8 & 56.7 & \textbf{56.1} & \textbf{50.7} & - \\
		\textbf{Predator\cite{huang2020predator} + \Name}     & \textbf{61.5} & \textbf{60.2} & \textbf{58.5} & 55.4 & 50.4 & +2.50 \\
        \Xhline{1.5\arrayrulewidth}
        \end{tabular}
    }
    
    \caption{Registration recall on the 3DLoMatch dataset using different numbers of points to construct the input correspondence set. The last column is the average increase brought by \Name.}
    \label{tab:3DLoMatch}
\end{table}

As shown in Table \ref{tab:3DLoMatch}, our method consistently outperforms RANSAC when combined with different descriptors. Moreover, our method can further boost the performance of Predator~\cite{huang2020predator}, a recently proposed learning-based descriptors especially designed for low-overlapping registration, showing the effectiveness and robustness of our method under high outlier ratios. \Name~increases the registration recall by \textbf{16.3\%} and \textbf{7.3\%} under 5000 points setting for FCGF and Predator, respectively. Note that \Name~does not bring much performance gain when only a small number of points~(e.g. less than 500) are used to construct the input correspondences mainly because some of the point cloud pairs have too few~(e.g. less than 3) correspondences to identify a unique registration.

\xy{
\noindent\textbf{Prioritized RANSAC.}
Despite the common usage of the inlier probability predicted by networks in weighted least-squares fitting~\cite{choy2020deep, gojcic2020LearningMultiview}, little attention has been drawn to leverage the predicted probability in a RANSAC framework. In this experiment, we derive a strong RANSAC variant~(denoted as \textit{Prioritized}) by using the inlier probability for selecting seeds to bias the sampling distribution. For a fair comparison, we implement \textit{Prioritized} using the same codebase~(\texttt{Open3D}) as RANSAC. As shown in Table~\ref{tab:abl_ransac}, \textit{Prioritized} outperforms classic RANSAC by more than 30\% in terms of registration recall, indicating that the inlier probability predicted by our method is meaningful and accurate, thus could help RANSAC to sample all-inlier subsets earlier and to achieve better performance in fewer iterations. This RANSAC variant can also be used for each correspondence subset to replace the weighted LS in Eq.~\ref{eq:solve_r_t}, denoted as \textit{Local Prioritized} in Table~\ref{tab:abl_ransac}. \Name~still outperforms the strong baselines with better accuracy and faster speed.}
\begin{table}[h]
    \centering
    \vspace{-0.2cm}
    \setlength{\abovecaptionskip}{0.10cm}
 	\setlength{\belowcaptionskip}{-0.20cm}
    \resizebox{0.45\textwidth}{!}{
        \begin{tabular}{l|cccccc}
        \Xhline{1.5\arrayrulewidth}
         & {RR(\%)} & {RE(\textdegree)} & {TE(cm)} & {F1(\%)} & {Time(s)}\\
        \hline
        
         \textbf{RANSAC-1k} & 	40.05 & 5.16 & 13.65 & 39.23 & 0.08  \\
		\textbf{Prioritized-1k} & 74.31 & 2.83 & 8.26 & 67.58 & 0.13\\
		\textbf{Local Prioritized} & 78.00 & 2.08 & \textbf{6.42} & 69.44 & 0.24\\
		\textbf{\Name} & \textbf{78.50} & \textbf{2.07} & 6.57 & \textbf{69.85} & 0.09 \\

        \Xhline{1.5\arrayrulewidth}
        \end{tabular}
    }
    \caption{Results on 3DMatch test set using FPFH. 
    }
    \label{tab:abl_ransac}
\end{table}

\noindent\textbf{Ablation on loss function.} The $L_{sm}$ is proposed to provide additional supervision, i.e., the pairwise relations between correspondences, serving as a complement to the node-wise supervision. The edge-wise supervision encourages the inliers to be concentrated in the embedding space, and this is the key assumption of our \nsm~module. To demonstrate its effectiveness, we compare the model {trained with Eq.~\ref{eq:full_loss}} and the model trained without the proposed spectral matching loss $L_{sm}$~(Eq.~\ref{eq:sm_loss}) in Table~\ref{tab:ablation_sm}. As shown in Table~\ref{tab:ablation_loss}, $L_{sm}$ improves the registration recall by 0.67\% over the strong baseline.

\begin{table}[h]
    \centering
     \setlength{\abovecaptionskip}{0.10cm}
    \setlength{\belowcaptionskip}{-0.20cm}
    \resizebox{0.4\textwidth}{!}{
        \begin{tabular}{l|ccccc}
        \Xhline{1.5\arrayrulewidth}
    
         & {RR($\uparrow$)} & {RE($\downarrow$)} & {TE($\downarrow$)} & {F1($\uparrow$)} & {Time}  \\
        \hline
         \textbf{\Name}     & \textbf{93.28} & \textbf{2.06} & \textbf{6.55} & \textbf{82.35} & 0.09 \\
         \textbf{~w/o $L_{sm}$}  & 92.61 & 2.07 & 6.75 & 81.58 & 0.09 \\

        \Xhline{1.5\arrayrulewidth}
        \end{tabular}
    }
    
    \caption{Ablation experiments of \nsm~module.} 
    \label{tab:ablation_loss}
\end{table}

\noindent\textbf{Effect of neighborhood size $k$.} The size of correspondence subset, $k$, (Sec.~\ref{subsec:nsm}) is a key parameter of our proposed method, {and} 
controls the size of each {correspondence} 
subset for neural spectral matching. We test the performance of our method with $k$ being 10, 20, 30, 40, 50, 60, 100, and 200, respectively. As shown in Table~\ref{tab:ablation_k}, the results show that our method is robust to the choice of $k$. We ascribe the robustness to the neural spectral matching module, which effectively prunes the potential outliers in the retrieved subsets, thus producing a correct model even when starting from a not-all-inlier sample. We finally choose $k=40$ for its best \textit{Registration Recall} and modest computation cost.

\begin{table}[h]
    \centering
     \setlength{\abovecaptionskip}{0.10cm}
    \setlength{\belowcaptionskip}{-0.20cm}
    \resizebox{0.40\textwidth}{!}{
        \begin{tabular}{l|cccccccc}
        \Xhline{1.5\arrayrulewidth}
         & {RR($\uparrow$)} & {RE($\downarrow$)} &  {TE($\downarrow$)}& {IP($\uparrow$)} & {IR($\uparrow$)} & {F1($\uparrow$)}  \\
        \hline
        \textbf{10}     & 92.73 & 2.04 & \textbf{6.44} & 79.01 & 85.51 & 81.87 \\
        \textbf{20}     & 92.79 & 2.04 & 6.50 & 78.88 & 85.86 & 81.96 \\
        \textbf{30}     & 93.10 & 2.04 & 6.50 & 79.07 & 86.35 & 82.25 \\
        \textbf{40}     & \textbf{93.28} & 2.06 & 6.55 & 79.10 & 86.54 & 82.35\\
        \textbf{50}     & 93.10 & 2.05 & 6.54 & 79.10 & 86.47 & 82.34 \\
        \textbf{60}     & 92.91 & 2.04 & 6.51 & \textbf{79.14} & \textbf{86.61} & \textbf{82.42} \\
        \textbf{100} 	& 92.91 & 2.04 & 6.53 & 78.87 & 86.25 & 82.12 \\
        \textbf{200}    & 92.79 & 2.04 & 6.51 & 78.96 & 86.37 & 82.22 \\

        \Xhline{1.5\arrayrulewidth}
        \end{tabular}
    }
    
    \caption{Performance of our \Name~when varying the size of correspondence subsets in the NSM module.}
    \label{tab:ablation_k}
\end{table}

\noindent\textbf{Joint training with descriptor and detector.}
In this part, we explore the potential of jointly optimizing the local feature learning and 
outlier rejection stages. A recently proposed method D3Feat \cite{bai2020d3feat}, which efficiently performs dense feature detection and description by a single network, best suits our need.
By back-propagating gradients to the input descriptors, the detector network can also be updated. Thus we build an end-to-end registration pipeline by taking the output of D3Feat as the input to our outlier rejection algorithm. We establish the correspondences using soft nearest neighbor search proposed in~\cite{gojcic2020LearningMultiview} to make the whole pipeline differentiable. We first train the feature network and the outlier rejection network separately, and then fine-tune them together using the losses in~\cite{bai2020d3feat} and Eq.~\ref{eq:full_loss}.
	
{However,} we did not observe performance improvement for the feature network in this preliminary joint training experiment. {We suspect that} 
the current losses are unable to provide meaningful gradients to the feature network. 
We believe that it is an interesting future direction 
to design proper loss formulations for end-to-end learning of both feature and outlier rejection networks. 

Nevertheless, it is noteworthy that within a reasonable range, D3Feat + \Name~achieves improved results when using {fewer} 
but more confident keypoints to build the input putative correspondences for outlier rejection. We ascribe the performance improvement to the elimination of keypoints in non-salient regions like smooth surface {regions}, 
{reducing} the failure registration caused by large symmetric objects in the scene. (See the visualization of failure cases Fig.~\ref{fig:failure} for more detail.) The results of D3Feat + \Name~under different numbers of keypoints~(labelled by \textbf{Joint(\#num)}) are provided in Table~\ref{tab:joint} for comparisons.

 \begin{table}[h]
     \centering
    \setlength{\abovecaptionskip}{0.10cm}
    \setlength{\belowcaptionskip}{-0.20cm}
     \resizebox{0.45\textwidth}{!}{
         \begin{tabular}{l|cccccc}
         \Xhline{2\arrayrulewidth}
          & {RR($\uparrow$)} & {RE($\downarrow$)} & {TE($\downarrow$)} & IP($\uparrow$) & IR($\uparrow$) & {F1($\uparrow$)} \\
          \hline
          \textbf{\Name} & 93.28 & 2.06 & 6.55 & 79.10 & \textbf{86.54} & 82.35 \\
          \textbf{Joint~(5000)} & 92.42 & \textbf{1.83} & \textbf{5.87} & 79.02 & 85.14 & 81.72  \\
          \textbf{Joint~(4000)} & 92.67 & 1.86 & 5.88 & 79.67 & 85.54 & 82.26 \\
          \textbf{Joint~(3000)} & 93.35 & 1.85 & 5.92 & 80.78 & 86.26 & 83.19\\
          \textbf{Joint~(2500)} & \textbf{93.59} & {1.86} & 6.00 & 81.05 & 86.40 & \textbf{83.38}  \\
          \textbf{Joint~(2000)} & 93.53 & {1.85} & 6.02 & \textbf{81.14} & 86.11 & 83.30  \\
          \textbf{Joint~(1000)} & 90.82 & {1.96} & 6.38 & 78.75 & 83.41 & 80.64  \\
         \Xhline{2\arrayrulewidth}
         \end{tabular}
     }
     \caption{Registration results of joint training with descriptor and detector on 3DMatch.}
     \label{tab:joint}
 \end{table}

\subsection{Derivation of Eq.~\ref{eq:solve_r_t}}
\label{supp:proof}
For completeness,
 we summarize the closed-form solution of the weighted least-squares pairwise registration problem~\cite{besl1992method},
\begin{equation}
    \mathbf{\hat R, \hat t} = \arg\min_{\mathbf{R, t}} \sum_{i}^{N} e_i\left\Vert \mathbf{R}\bm{x_i} + \mathbf{t} - \bm{y_i}  \right\Vert^2,
\end{equation}
{where $\bm{(x_i, y_i)}$ is a pair of corresponding points, with $\bm{x_i}$ and $\bm{y_i}$ being from point clouds $\mathbf{X} \in \mathbb{R}^{N\times 3}$ and $\mathbf{Y}\in \mathbb{R}^{N\times 3}$, respectively.} 
Let $\bm{\bar x}$ and $\bm{\bar y}$ denote the weighted centroids of {$\mathbf{X}$ and $\mathbf{Y}$, respectively}: 
\begin{equation}
	\bm{\bar x} = \frac{\sum_{i}^{N}e_i\bm{x_i}}{\sum_{i}^N e_i}, \ \ \ \bm{\bar y} = \frac{\sum_{i}^{N}e_i\bm{y_i}}{\sum_{i}^N e_i}.
\end{equation}
We first convert the original coordinates to {the} centered coordinates by subtracting the {corresponding} centroids, 
\begin{equation}
	\bm{\tilde x_i = x_i - \bar x, \ \ \tilde y_i = y_i - \bar y},~~~ i = 1,...,N.
\end{equation}
The next step involves {the calculation of} 
the weighted covariance matrix $\mathbf{H}$, 
\begin{equation}
	\mathbf{H = \tilde X^TE\tilde Y},
\end{equation}
where $\mathbf{\tilde X}$ and $\mathbf{\tilde Y}$ are the matrix forms of {the} centered coordinates and $\mathbf{E}=\text{diag}(e_1, e_2, ..., e_N)$. Then the rotation matrix from $\mathbf{X}$ to $\mathbf{Y}$ can be found by singular value decomposition (SVD),
\begin{equation}
	\mathbf{[U,S,V] = \text{SVD}(H)},
\end{equation}
\begin{equation}
\mathbf{
	\hat R = V \left[               
  \begin{array}{ccc}   
    1 & 0 & 0\\  
    0 & 1 & 0\\ 
    0 & 0 & \text{det}(\mathbf{VU^T})
  \end{array}
   \right]
U^T ,}
\end{equation}
where $\text{det}(\cdot)$ denotes the determinant, which is used to avoid
 the reflection cases. Finally, the translation between the two point clouds is computed as,
\begin{equation}
	\mathbf{\hat t} = \bm{\bar y} - \mathbf{\hat R} \bm{\bar x}.
\end{equation}

\subsection{Qualitative Results}
\label{supp:vis}
\begin{figure*}[b]
    \includegraphics[width=9cm]{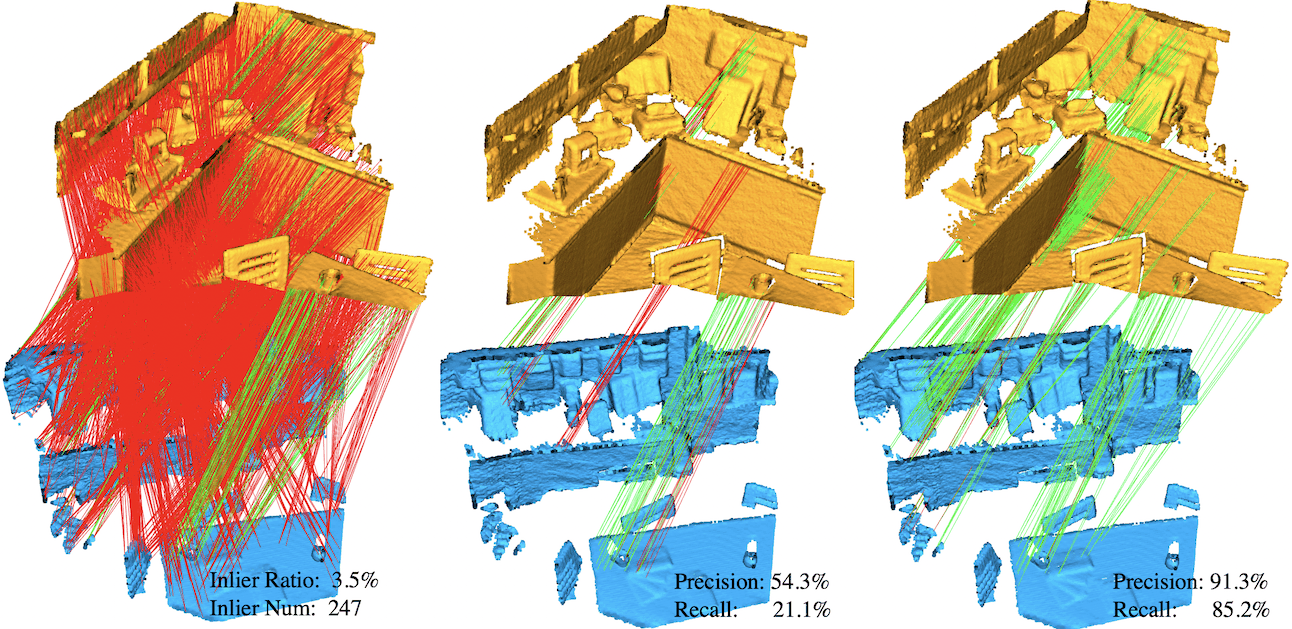}\hspace{0.5cm}\vspace{0.5cm}
    \includegraphics[width=8.7cm]{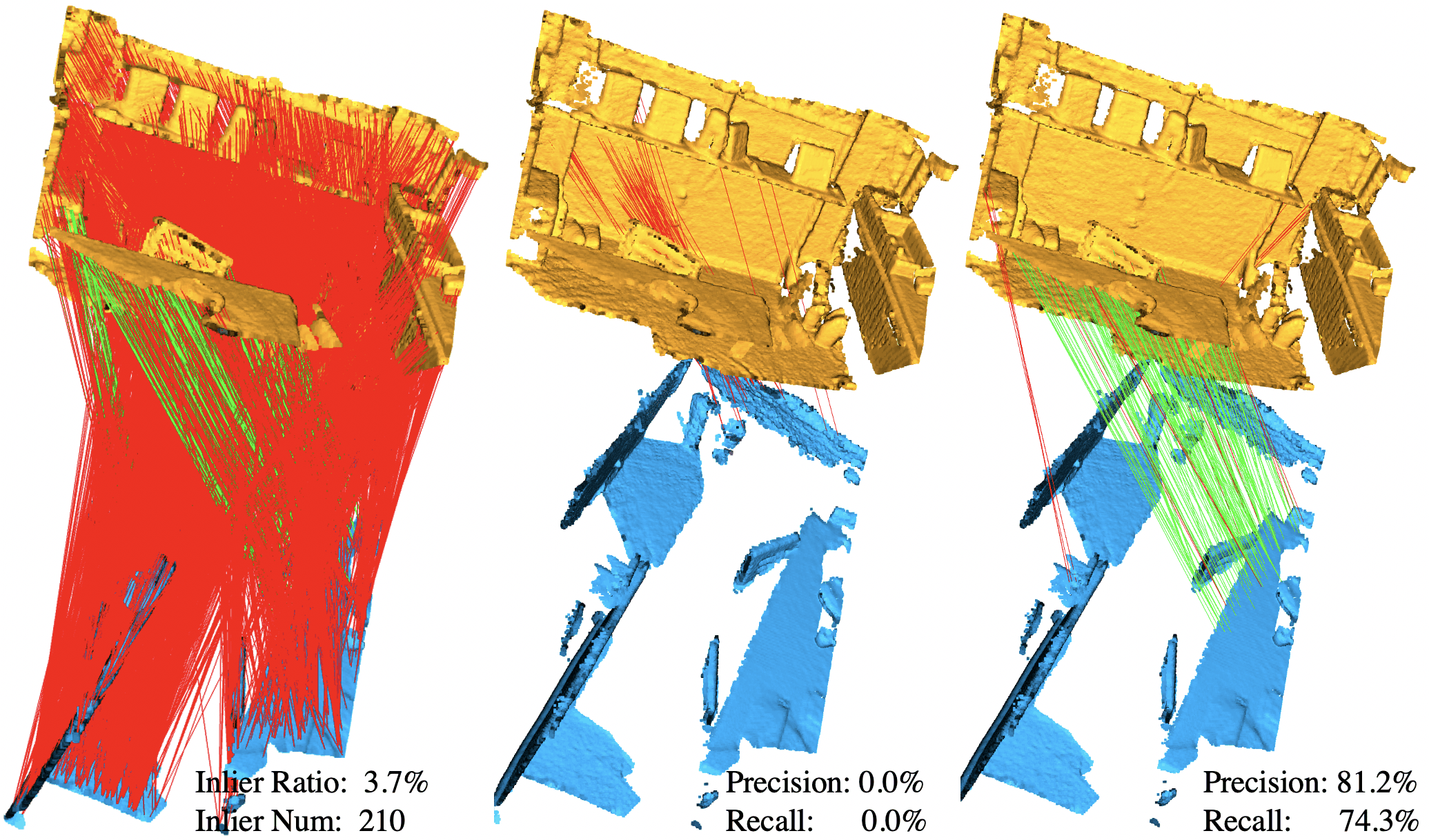} 
    
    \includegraphics[width=9cm]{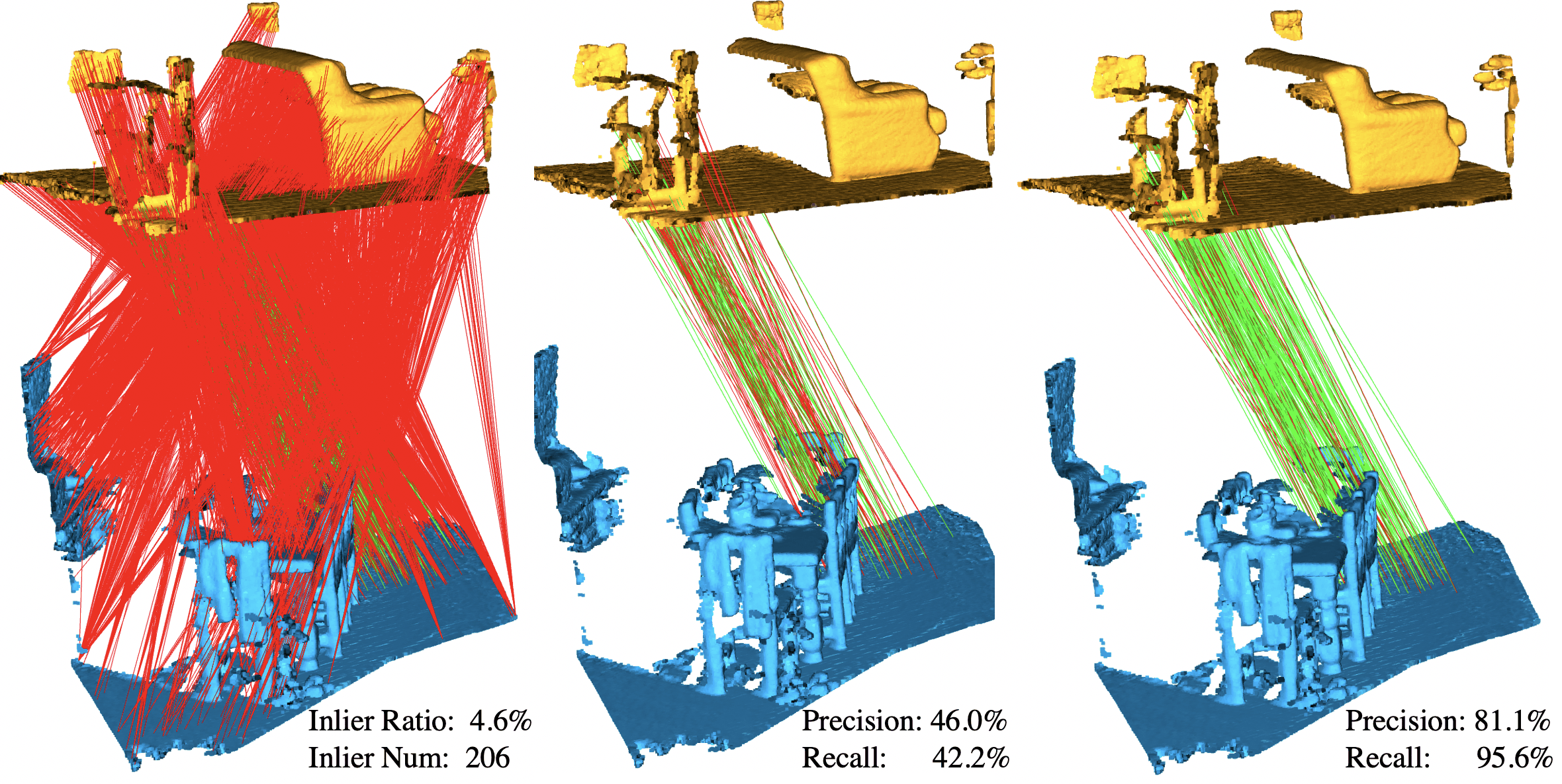}\hspace{0.5cm}\vspace{0.5cm}
    \includegraphics[width=8.7cm]{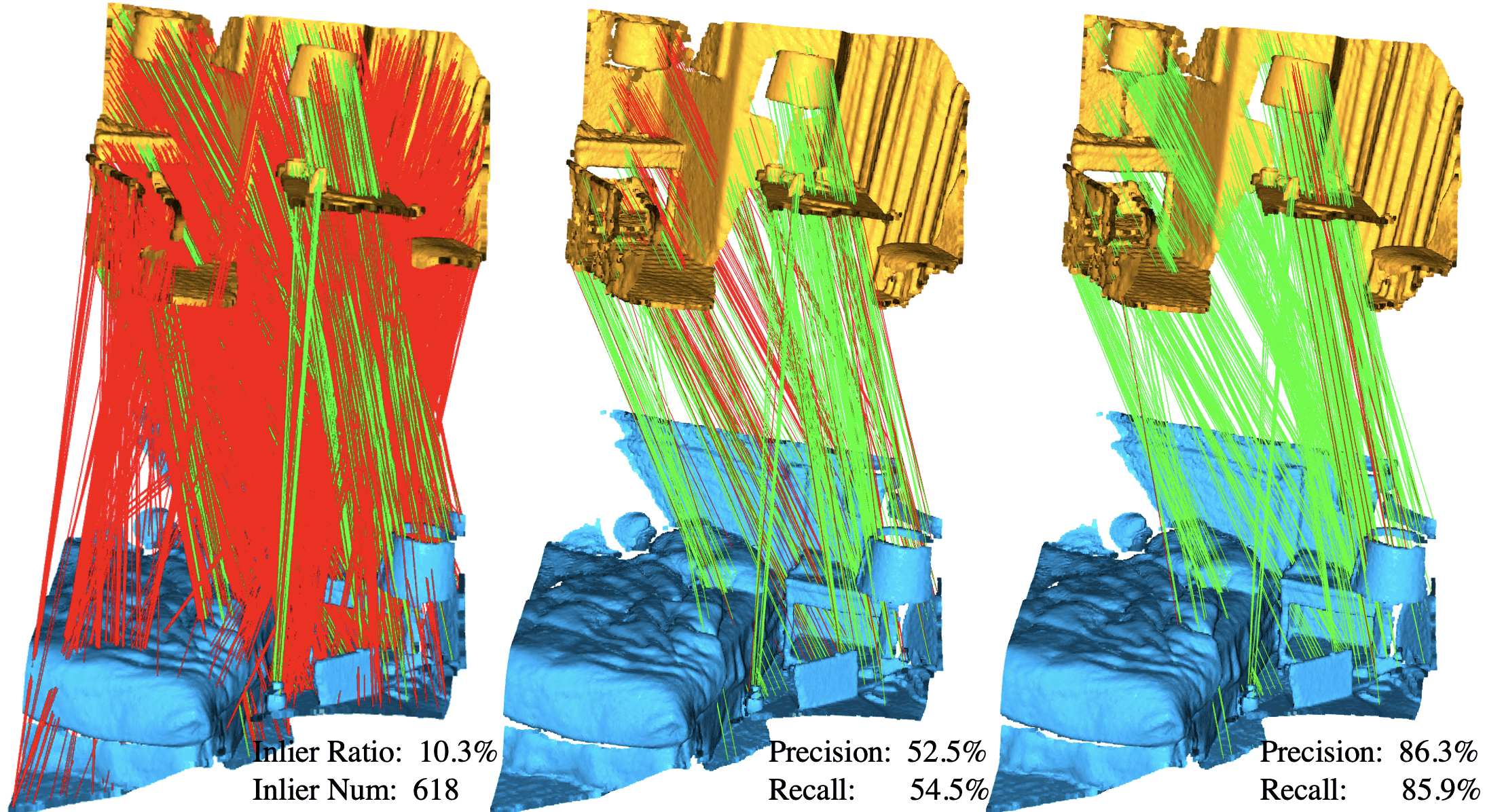}
    
    \includegraphics[width=9cm]{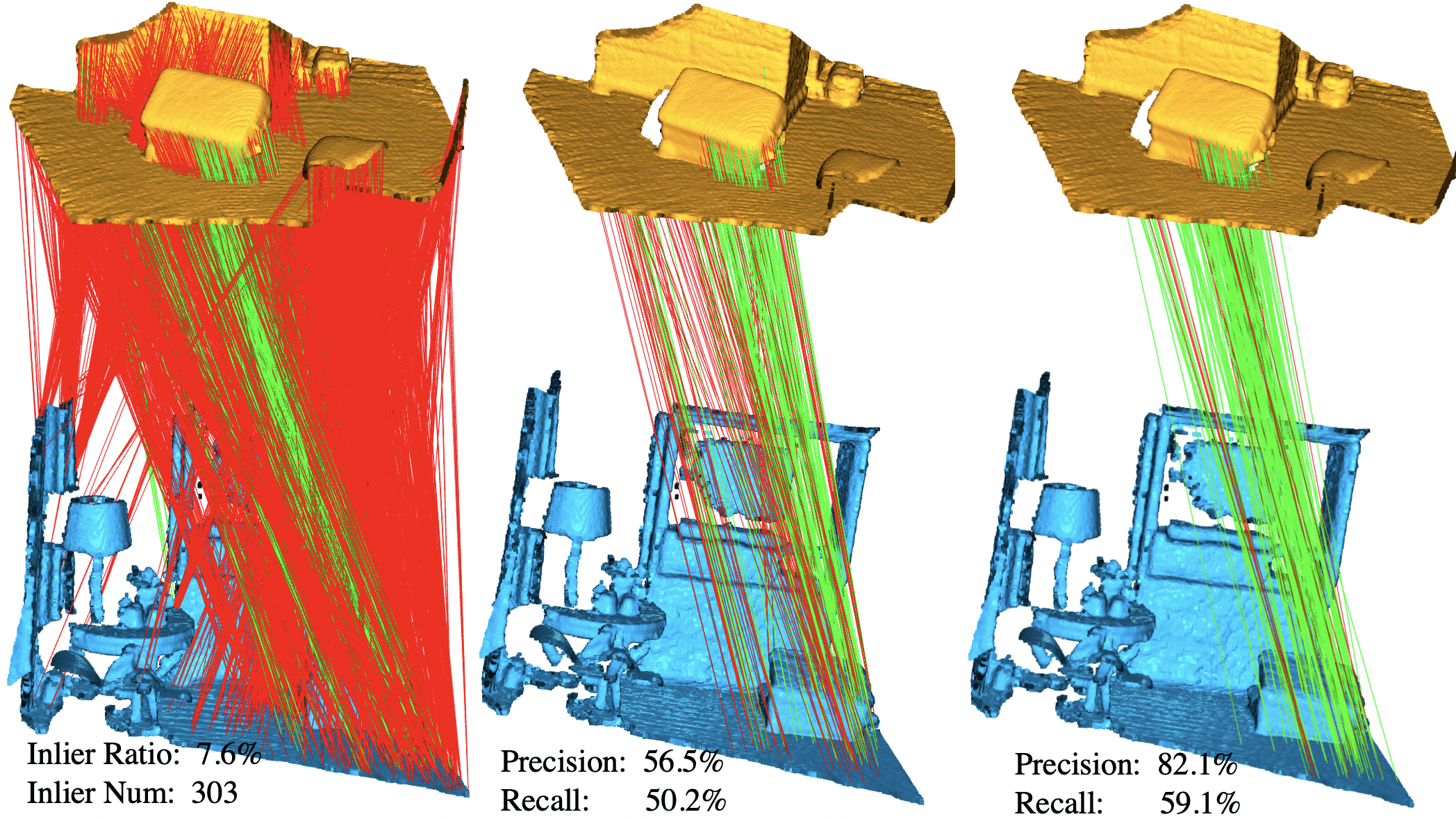}\hspace{0.5cm}\vspace{0.5cm}
    \includegraphics[width=9cm]{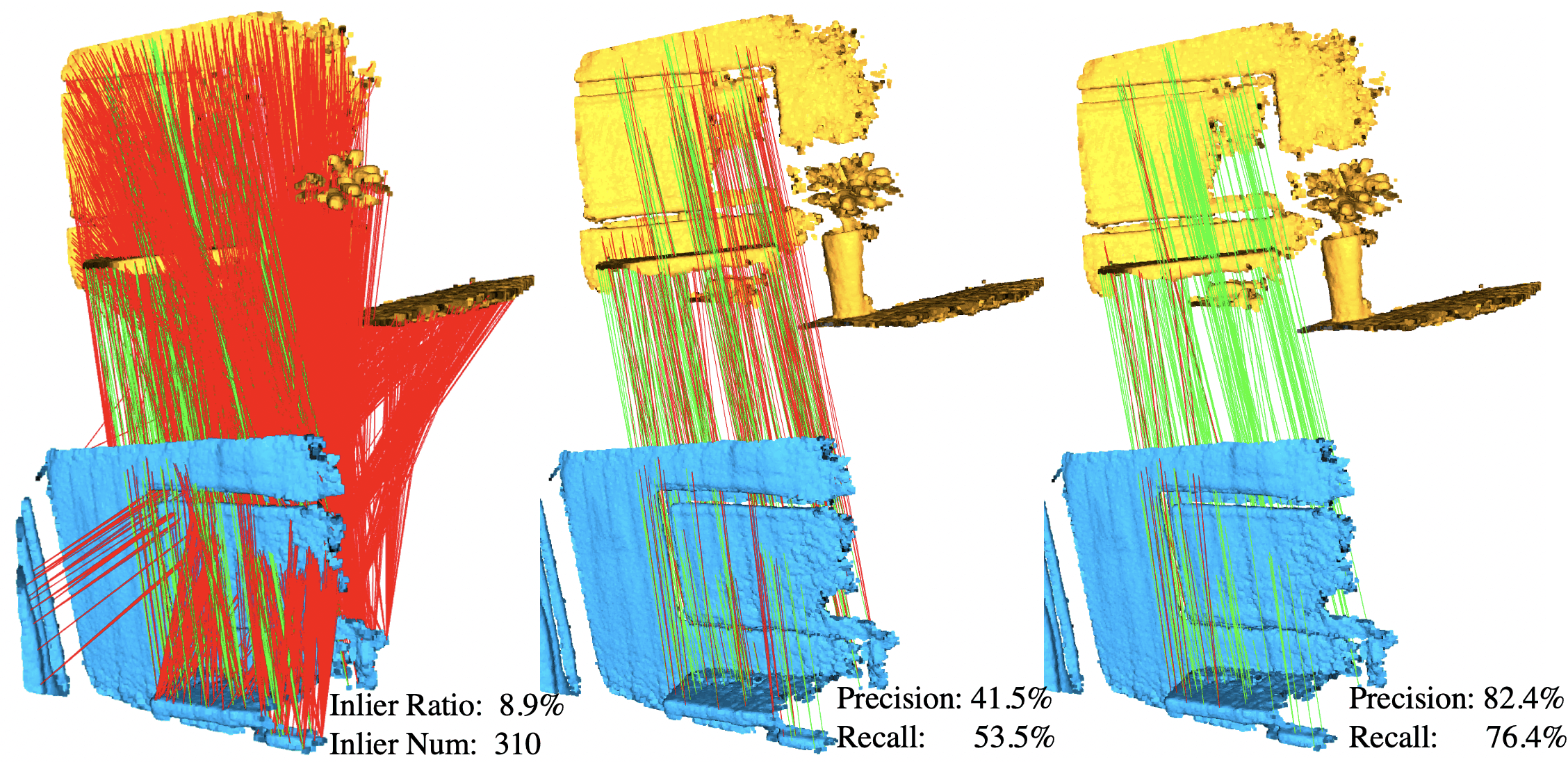}
    
    \includegraphics[width=8.9cm]{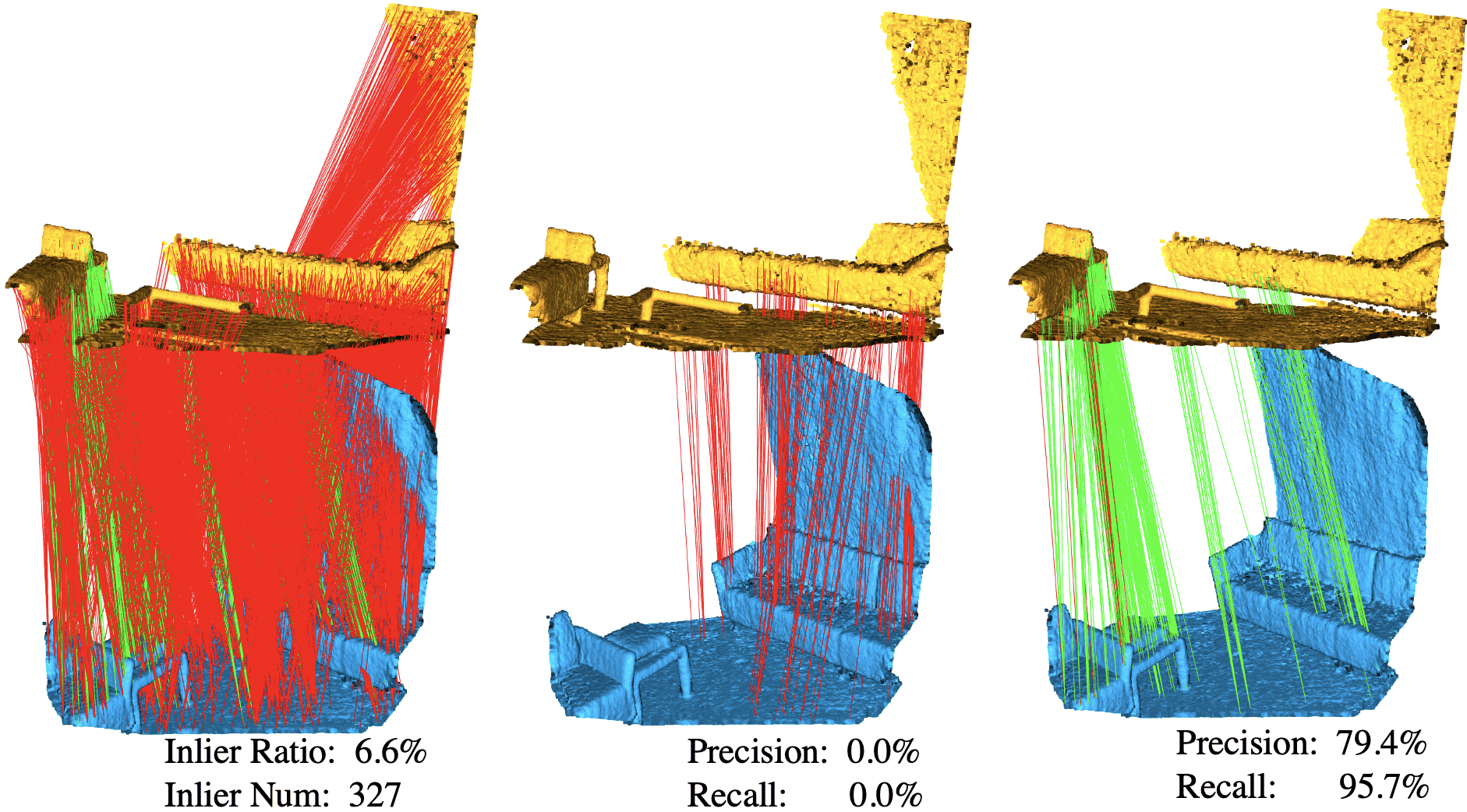}\hspace{0.6cm}
    \includegraphics[width=8.9cm]{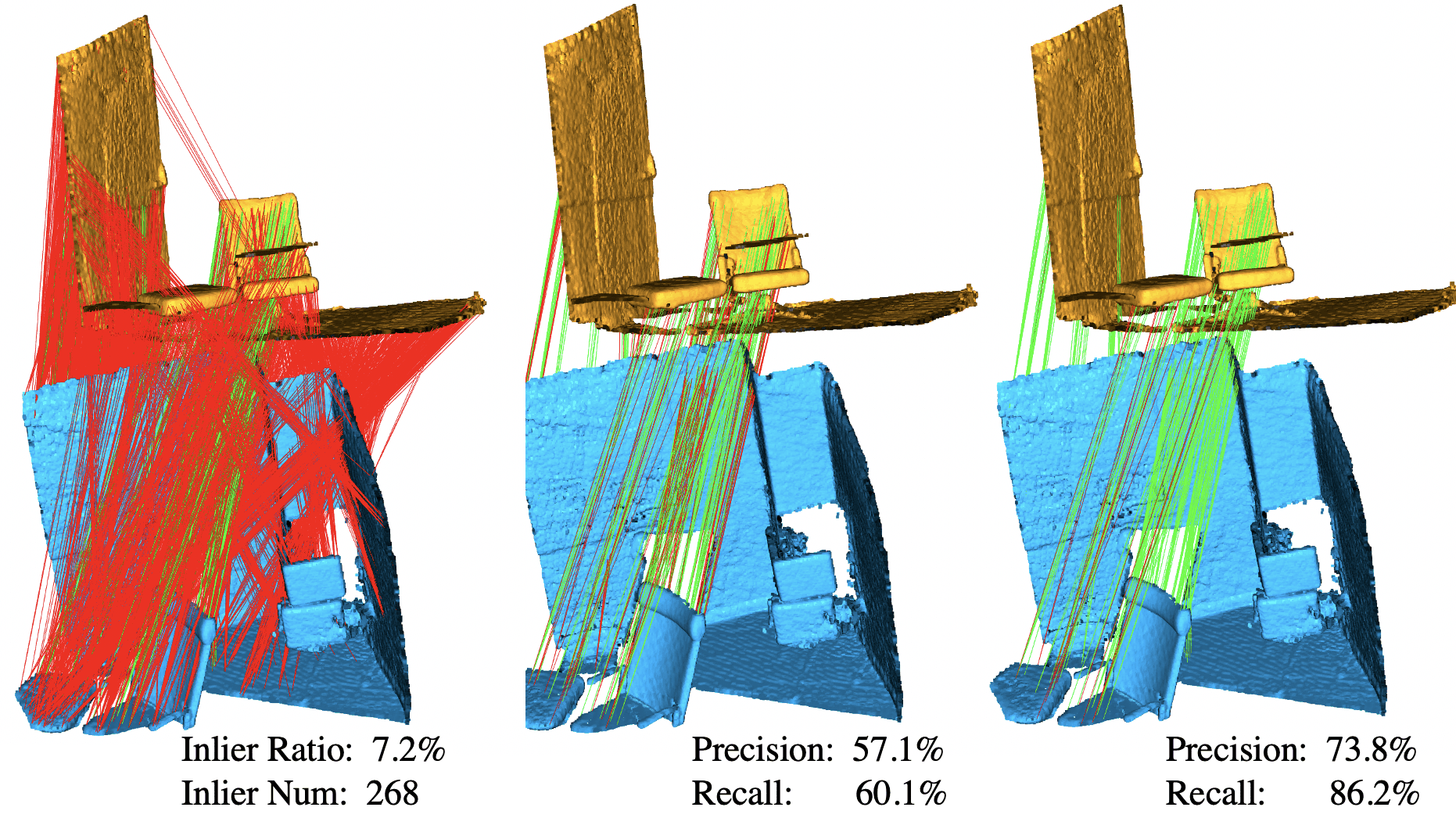}
     \caption{Visualization of outlier rejection results on the 3DMatch dataset. From left to right: input correspondences constructed by FCGF, results of RANSAC-100k, and results of \Name. Best viewed with color and zoom-in.}
    \label{fig:vis_3dmatch}
\end{figure*}

\begin{figure*}[t]
    \centering  
    \includegraphics[width=8.5cm]{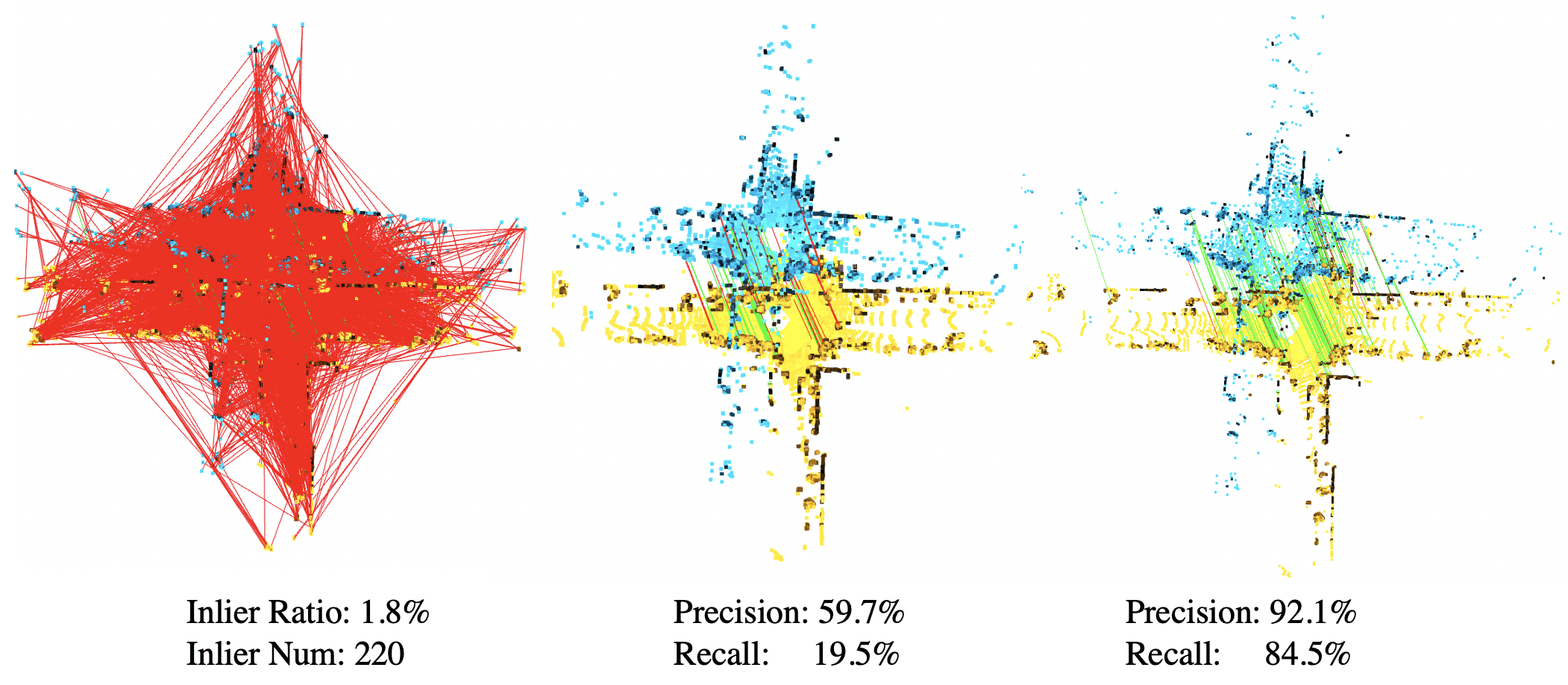}
    \includegraphics[width=8.5cm]{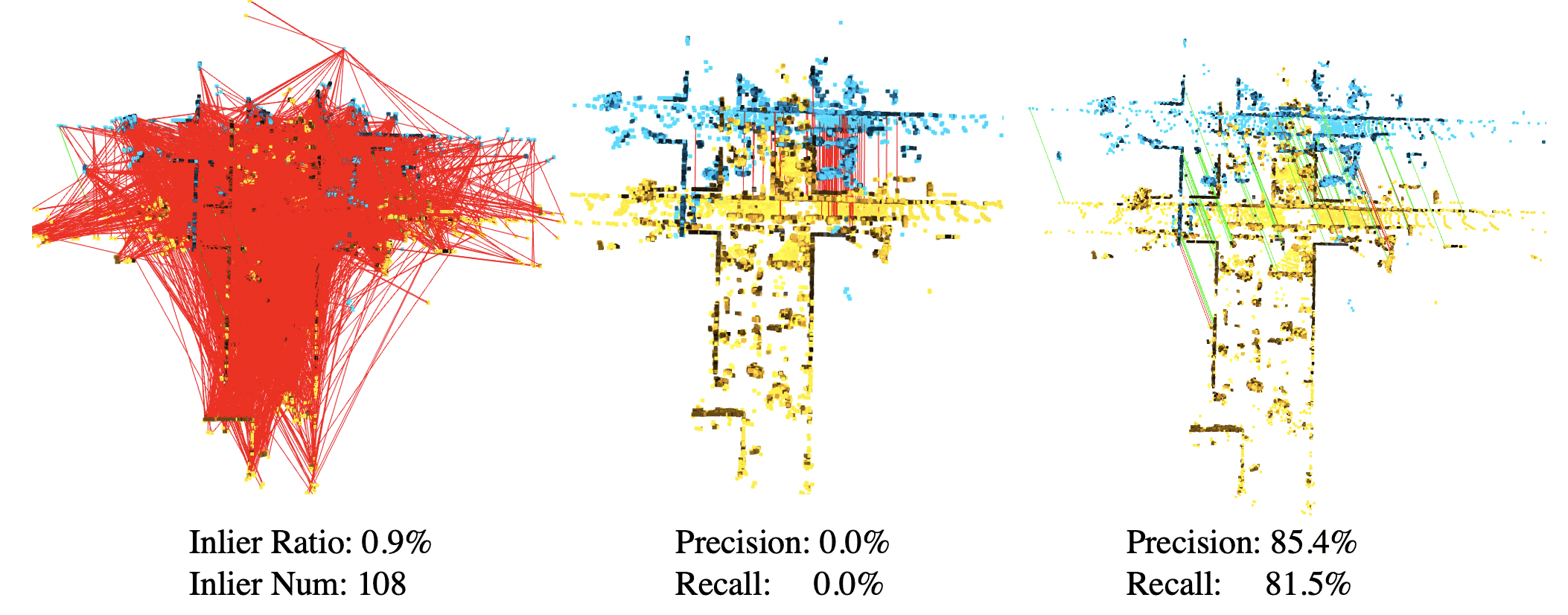} 
    
    \includegraphics[width=8.5cm]{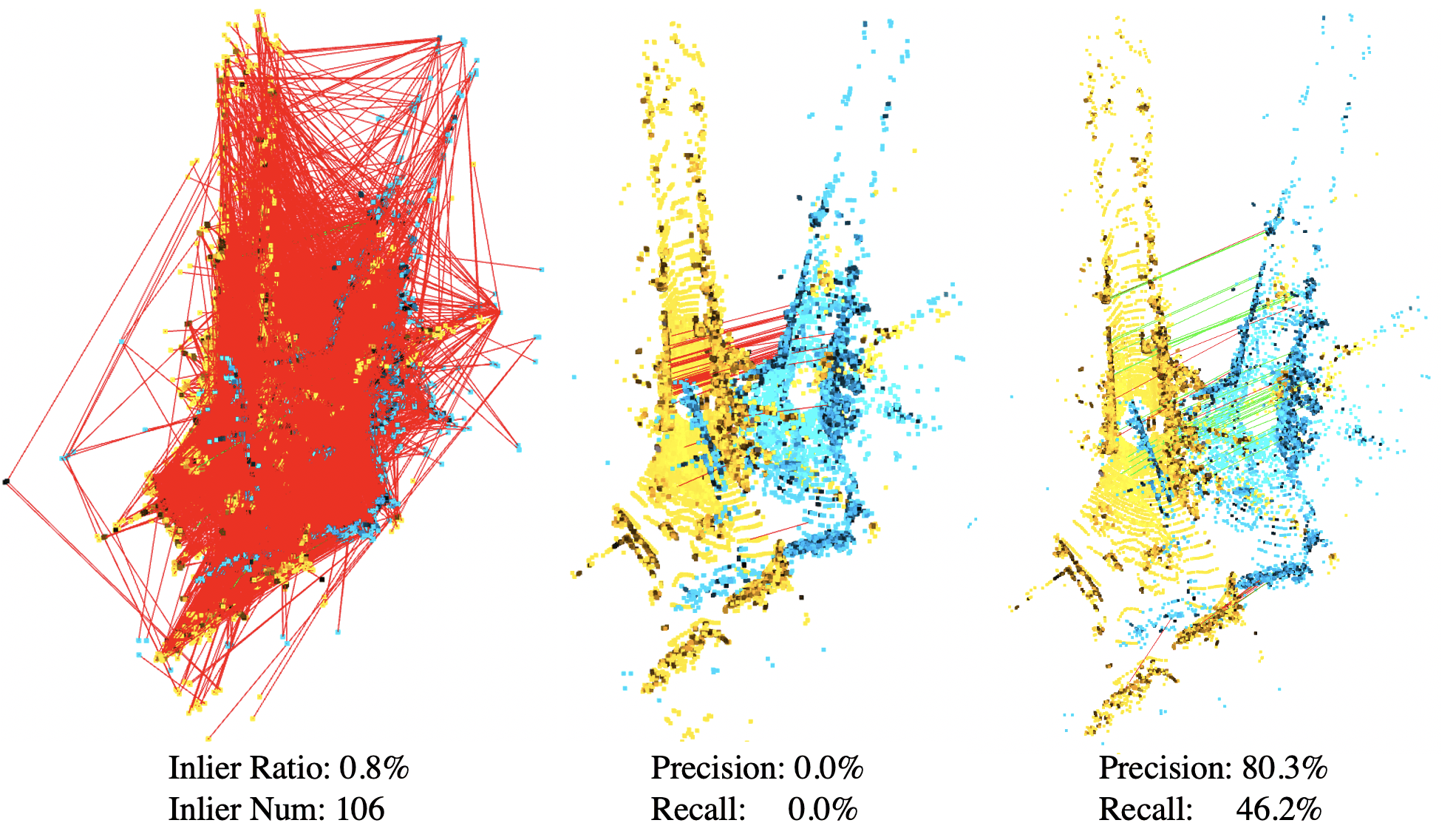}
    \includegraphics[width=8.5cm]{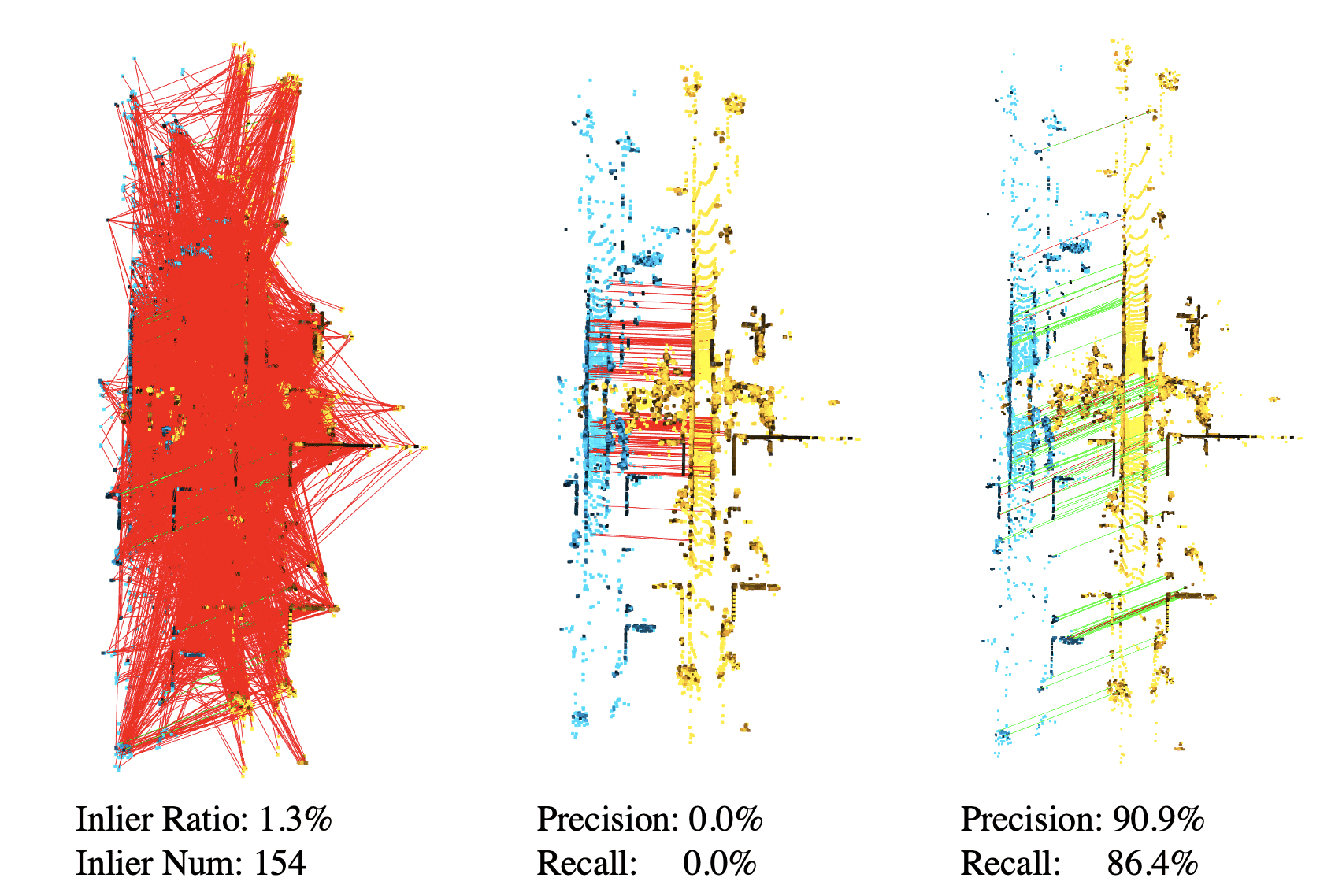}
    	
    \includegraphics[width=8.5cm]{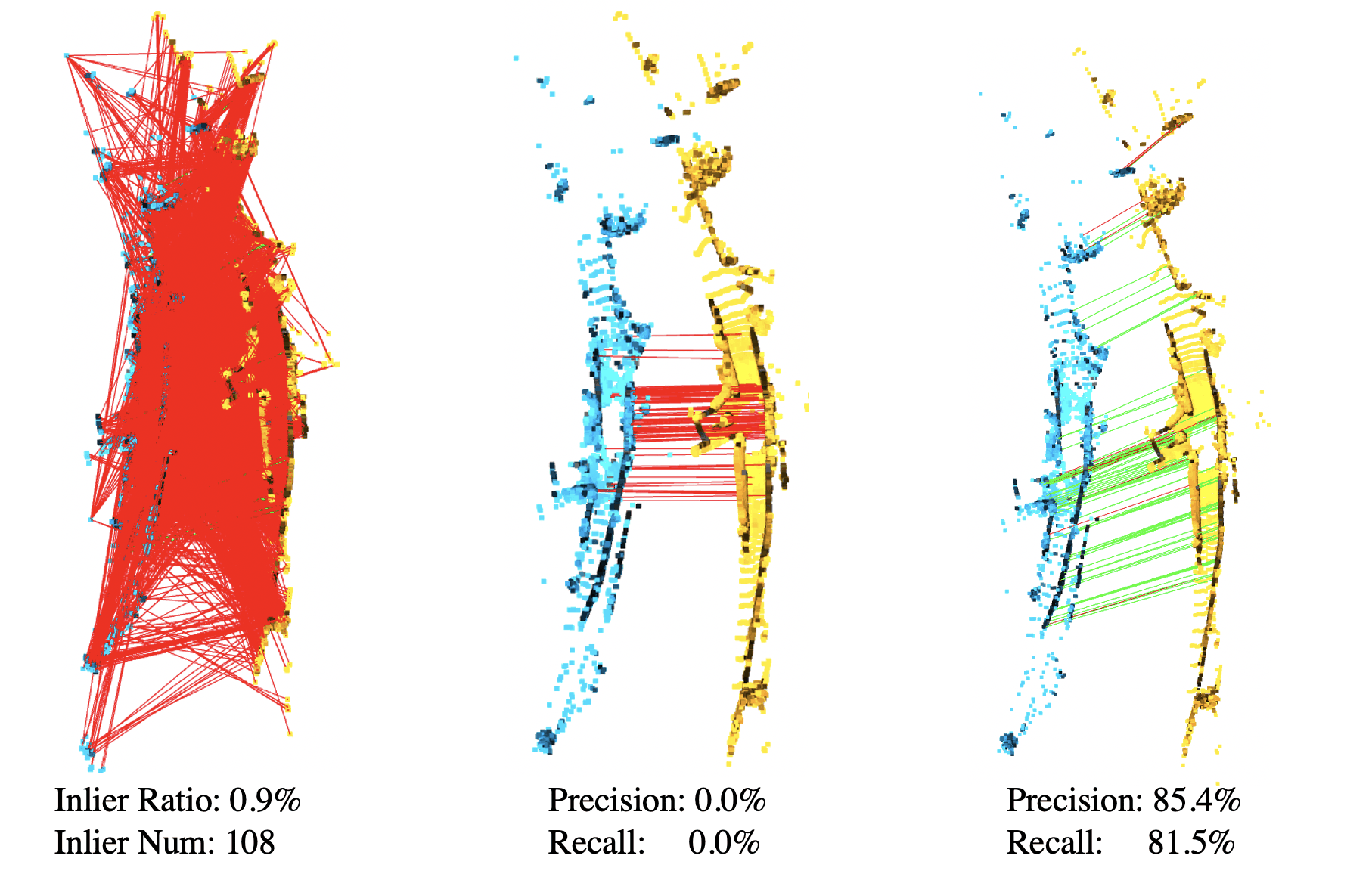}
    \includegraphics[width=8.5cm]{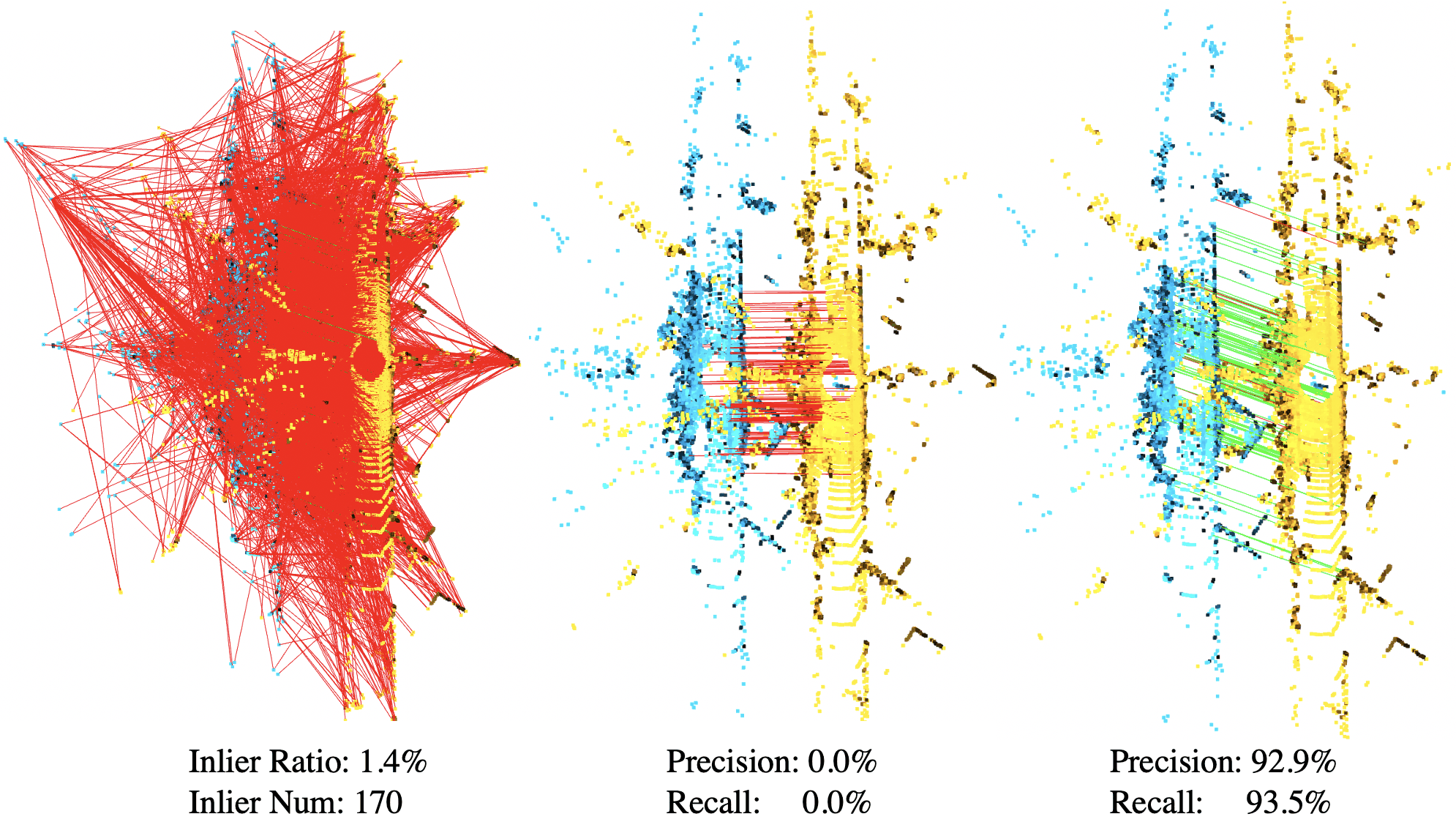}
    
     \caption{Visualization of outlier rejection results on the KITTI dataset. From left to right: input correspondences constructed by FPFH~(we choose FPFH to better demonstrate the robustness of our method to high outlier ratios), results of RANSAC-100k, and results of \Name. Best viewed with color and zoom-in.}
    \label{fig:vis_kitti}
\end{figure*}

We show the outlier rejection results on 3DMatch and KITTI in Fig.~\ref{fig:vis_3dmatch} and Fig.~\ref{fig:vis_kitti}, respectively. For the KITTI dataset, we use the FPFH descriptor to better demonstrate the superiority of our method. RANSAC suffers from significant performance degradation because the FPFH descriptor results in large outlier ratios, where it is harder to sample an outlier-free set. In contrast, 
our \Name~still gives satisfactory results.

\begin{figure}[h]
    \centering  
     \setlength{\abovecaptionskip}{0.10cm}
    \setlength{\belowcaptionskip}{-0.20cm}
     \includegraphics[width=8.5cm]{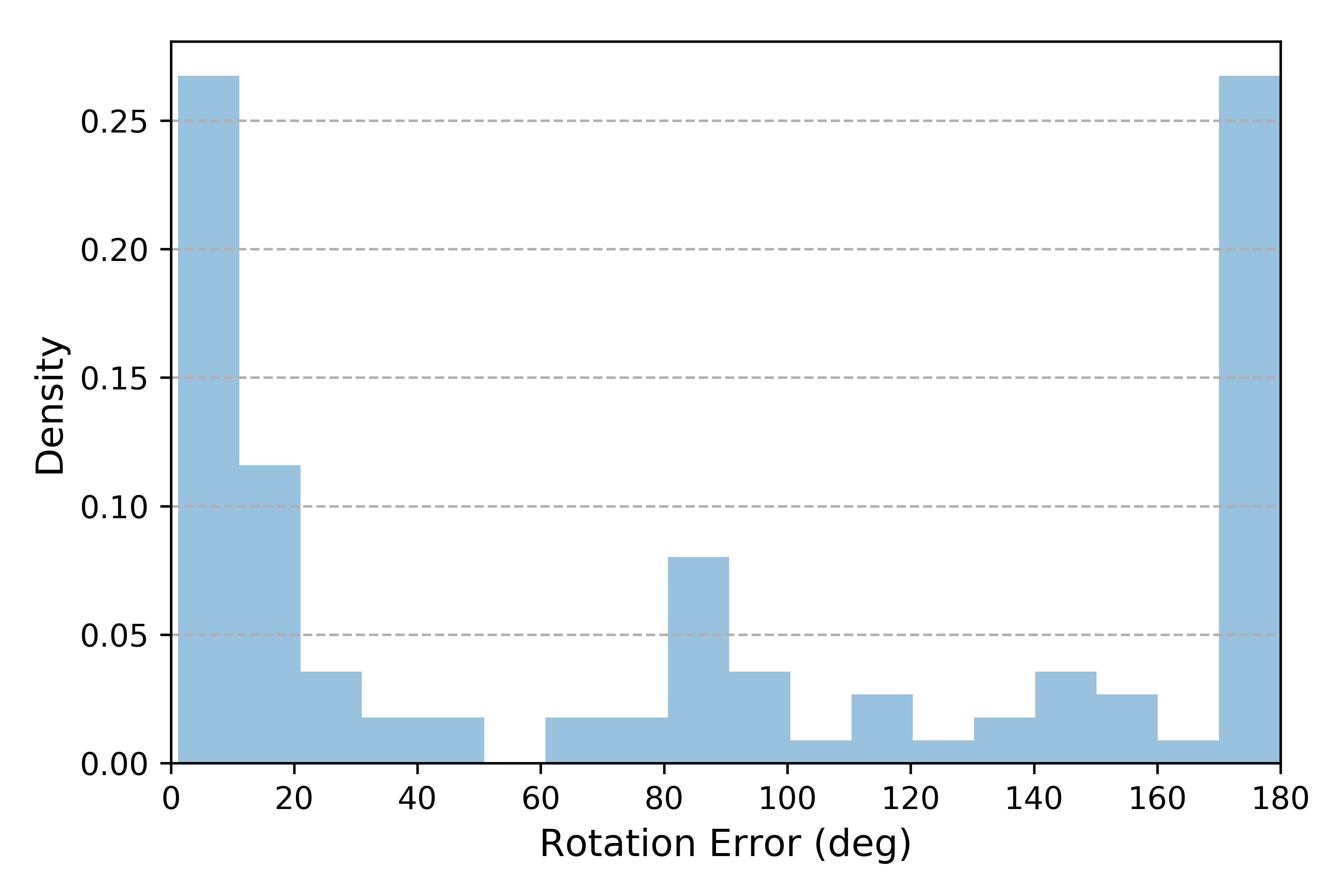}
     \caption{Rotation errors of unsuccessful registration pairs of \Name~on the 3DMatch test set.}
    \label{fig:re}
\end{figure}

\begin{figure*}[b]
    \centering  
	 \includegraphics[width=8.cm]{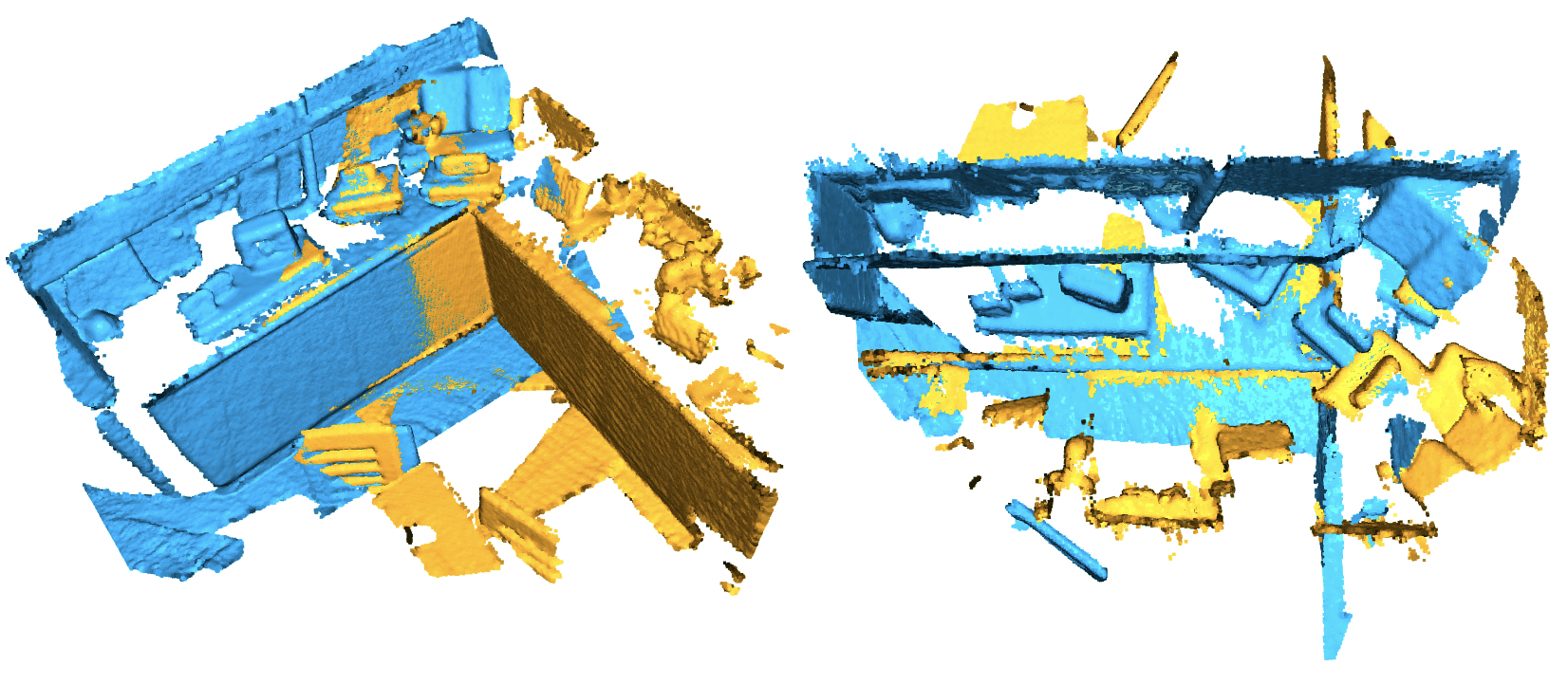}\hspace{1.0cm}
    \includegraphics[width=8.cm]{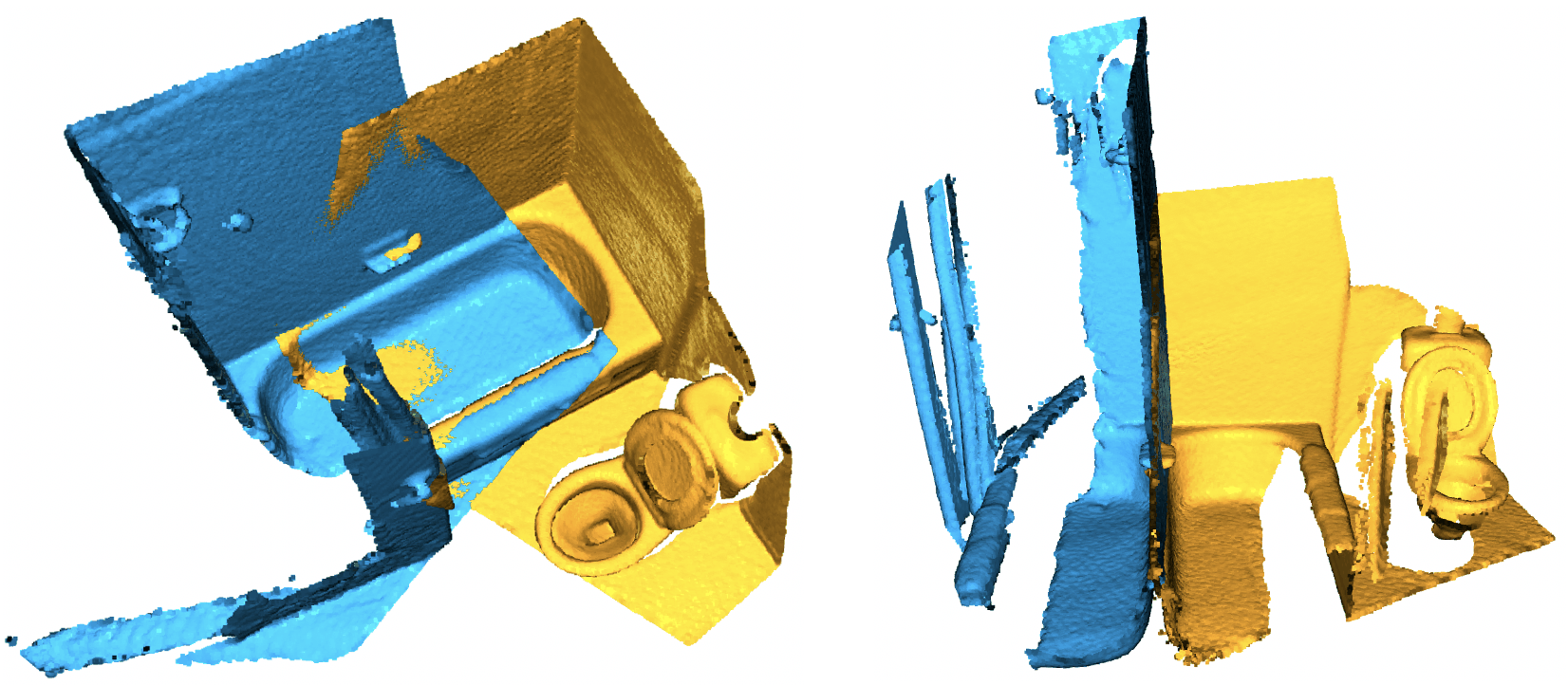} 
     \caption{Two representative failure examples 
     of our method on 3DMatch. In each example, ground-truth registration (Left) and estimated registration (Right). {We observe that} our method fails {mainly} due to the symmetries in the scene. 
     }
    \label{fig:failure}
\end{figure*}

We also provide {the} visualization of failure cases {of our method} on 3DMatch in Fig.~\ref{fig:failure}. One common failure {case happens} 
when there are large symmetry objects~(e.g., the wall, floor) in a scene, resulting in rotation errors around 90\textdegree or 180\textdegree. In {this case}, 
the clusters formed by outlier correspondences could become dominant, leading to incorrect transformation hypotheses. Then an incorrect transformation is probably selected as the final solution since a large number of outlier correspondences would satisfy this transformation. To highlight this issue, we draw the distribution of rotation errors of unsuccessful registration pairs on the 3DMatch test set in Fig.~\ref{fig:re}, from which we can find that a large portion of pairs has around 90\textdegree and 180\textdegree. 

%% file: egpaper_for_review.bbl
\begin{thebibliography}{10}\itemsep=-1pt

\bibitem{aiger20084}
Dror Aiger, Niloy~J Mitra, and Daniel Cohen-Or.
\newblock 4-points congruent sets for robust pairwise surface registration.
\newblock In {\em SIGGRAPH}. 2008.

\bibitem{ao2020SpinNet}
Sheng Ao, Qingyong Hu, Bo Yang, Andrew Markham, and Yulan Guo.
\newblock {SpinNet}: Learning a general surface descriptor for 3d point cloud
  registration.
\newblock {\em arXiv}, 2020.

\bibitem{yaoki2019pointnetlk}
Yasuhiro Aoki, Hunter Goforth, Rangaprasad Arun~Srivatsan, and Simon Lucey.
\newblock {PointNetLK}: Robust \& efficient point cloud registration using
  pointnet.
\newblock In {\em CVPR}, 2019.

\bibitem{bai2020d3feat}
Xuyang Bai, Zixin Luo, Lei Zhou, Hongbo Fu, Long Quan, and Chiew-Lan Tai.
\newblock {D3Feat}: Joint learning of dense detection and description of 3d
  local features.
\newblock In {\em CVPR}, 2020.

\bibitem{barath2018graph}
Daniel Barath and Ji{\v{r}}{\'\i} Matas.
\newblock Graph-cut ransac.
\newblock In {\em CVPR}, 2018.

\bibitem{bay2006surf}
Herbert Bay, Tinne Tuytelaars, and Luc Van~Gool.
\newblock {SURF}: Speeded up robust features.
\newblock In {\em ECCV}, 2006.

\bibitem{bergstrom2014robust}
Per Bergstr{\"o}m and Ove Edlund.
\newblock Robust registration of point sets using iteratively reweighted least
  squares.
\newblock {\em Computational Optimization and Applications}, 2014.

\bibitem{besl1992method}
Paul~J Besl and Neil~D McKay.
\newblock Method for registration of 3-d shapes.
\newblock In {\em Sensor fusion IV: control paradigms and data structures}.
  International Society for Optics and Photonics, 1992.

\bibitem{bian2017gms}
JiaWang Bian, Wen-Yan Lin, Yasuyuki Matsushita, Sai-Kit Yeung, Tan-Dat Nguyen,
  and Ming-Ming Cheng.
\newblock {GMS}: Grid-based motion statistics for fast, ultra-robust feature
  correspondence.
\newblock In {\em CVPR}, 2017.

\bibitem{opencv_library}
G. Bradski.
\newblock {The OpenCV Library}.
\newblock {\em Dr. Dobb's Journal of Software Tools}, 2000.

\bibitem{bustos2017guaranteed}
{\'A}lvaro~Parra Bustos and Tat-Jun Chin.
\newblock Guaranteed outlier removal for point cloud registration with
  correspondences.
\newblock {\em PAMI}, 2017.

\bibitem{bustos2019practical}
Alvaro~Parra Bustos, Tat-Jun Chin, Frank Neumann, Tobias Friedrich, and
  Maximilian Katzmann.
\newblock A practical maximum clique algorithm for matching with pairwise
  constraints.
\newblock {\em arXiv}, 2019.

\bibitem{cavalli2020adalam}
Luca Cavalli, Viktor Larsson, Martin~Ralf Oswald, Torsten Sattler, and Marc
  Pollefeys.
\newblock {AdaLAM}: Revisiting handcrafted outlier detection.
\newblock {\em arXiv}, 2020.

\bibitem{cho2010reweighted}
Minsu Cho, Jungmin Lee, and Kyoung~Mu Lee.
\newblock Reweighted random walks for graph matching.
\newblock In {\em ECCV}, 2010.

\bibitem{choi2015robust}
Sungjoon Choi, Qian-Yi Zhou, and Vladlen Koltun.
\newblock Robust reconstruction of indoor scenes.
\newblock In {\em CVPR}, 2015.

\bibitem{choy2020deep}
Christopher Choy, Wei Dong, and Vladlen Koltun.
\newblock Deep global registration.
\newblock In {\em CVPR}, 2020.

\bibitem{choy20194d}
Christopher Choy, JunYoung Gwak, and Silvio Savarese.
\newblock 4d spatio-temporal convnets: Minkowski convolutional neural networks.
\newblock In {\em CVPR}, 2019.

\bibitem{Choy_2019_ICCV}
Christopher Choy, Jaesik Park, and Vladlen Koltun.
\newblock Fully convolutional geometric features.
\newblock In {\em ICCV}, 2019.

\bibitem{chum2003locally}
Ond{\v{r}}ej Chum, Ji{\v{r}}{\'\i} Matas, and Josef Kittler.
\newblock Locally optimized ransac.
\newblock In {\em JPRS}. Springer, 2003.

\bibitem{cour2007balanced}
Timothee Cour, Praveen Srinivasan, and Jianbo Shi.
\newblock Balanced graph matching.
\newblock In {\em NeurIPS}, 2007.

\bibitem{deng2018ppf}
Haowen Deng, Tolga Birdal, and Slobodan Ilic.
\newblock {PPF-FoldNet}: Unsupervised learning of rotation invariant 3d local
  descriptors.
\newblock In {\em ECCV}, 2018.

\bibitem{deng2018ppfnet}
Haowen Deng, Tolga Birdal, and Slobodan Ilic.
\newblock {PPFNet}: Global context aware local features for robust 3d point
  matching.
\newblock In {\em CVPR}, 2018.

\bibitem{deng20193d}
Haowen Deng, Tolga Birdal, and Slobodan Ilic.
\newblock 3d local features for direct pairwise registration.
\newblock In {\em CVPR}, 2019.

\bibitem{fischler1981random}
Martin~A. {Fischler} and Robert~C. {Bolles}.
\newblock Random sample consensus: a paradigm for model fitting with
  applications to image analysis and automated cartography.
\newblock {\em Communications of The ACM}, 1981.

\bibitem{Geiger2013IJRR}
Andreas Geiger, Philip Lenz, Christoph Stiller, and Raquel Urtasun.
\newblock Vision meets robotics: The kitti dataset.
\newblock {\em IJRR}, 2013.

\bibitem{glent2014search}
Anders Glent~Buch, Yang Yang, Norbert Kruger, and Henrik Gordon~Petersen.
\newblock In search of inliers: 3d correspondence by local and global voting.
\newblock In {\em CVPR}, 2014.

\bibitem{gojcic2020LearningMultiview}
Zan Gojcic, Caifa Zhou, Jan~D Wegner, Leonidas~J Guibas, and Tolga Birdal.
\newblock Learning multiview 3d point cloud registration.
\newblock In {\em CVPR}, 2020.

\bibitem{gojcic2019perfect}
Zan Gojcic, Caifa Zhou, Jan~D Wegner, and Andreas Wieser.
\newblock The perfect match: 3d point cloud matching with smoothed densities.
\newblock In {\em CVPR}, 2019.

\bibitem{handa2014benchmark}
Ankur Handa, Thomas Whelan, John McDonald, and Andrew~J Davison.
\newblock A benchmark for rgb-d visual odometry, 3d reconstruction and slam.
\newblock In {\em ICRA}, 2014.

\bibitem{he2020pvn3d}
Yisheng He, Wei Sun, Haibin Huang, Jianran Liu, Haoqiang Fan, and Jian Sun.
\newblock {PVN3D}: A deep point-wise 3d keypoints voting network for 6dof pose
  estimation.
\newblock In {\em CVPR}, 2020.

\bibitem{holland1977robust}
Paul~W Holland and Roy~E Welsch.
\newblock Robust regression using iteratively reweighted least-squares.
\newblock {\em Communications in Statistics-theory and Methods}, 1977.

\bibitem{huang2020predator}
Shengyu Huang, Zan Gojcic, Mikhail Usvyatsov, Andreas Wieser, and Konrad
  Schindler.
\newblock {PREDATOR}: Registration of 3d point clouds with low overlap.
\newblock {\em arXiv}, 2020.

\bibitem{jian2010robust}
Bing Jian and Baba~C Vemuri.
\newblock Robust point set registration using gaussian mixture models.
\newblock {\em PAMI}, 2010.

\bibitem{jian20183dfeat}
Zi Jian~Yew and Gim Hee~Lee.
\newblock {3DFeat-Net}: Weakly supervised local 3d features for point cloud
  registration.
\newblock In {\em ECCV}, 2018.

\bibitem{kahler2016real}
Olaf K{\"a}hler, Victor~A Prisacariu, and David~W Murray.
\newblock Real-time large-scale dense 3d reconstruction with loop closure.
\newblock In {\em ECCV}, 2016.

\bibitem{kummerle2011g}
Rainer K{\"u}mmerle, Giorgio Grisetti, Hauke Strasdat, Kurt Konolige, and
  Wolfram Burgard.
\newblock g2o: A general framework for graph optimization.
\newblock In {\em ICRA}, 2011.

\bibitem{le2019sdrsac}
Huu~M Le, Thanh-Toan Do, Tuan Hoang, and Ngai-Man Cheung.
\newblock {SDRSAC}: Semidefinite-based randomized approach for robust point
  cloud registration without correspondences.
\newblock In {\em CVPR}, 2019.

\bibitem{leordeanu2005spectral}
Marius Leordeanu and Martial Hebert.
\newblock A spectral technique for correspondence problems using pairwise
  constraints.
\newblock In {\em ICCV}, 2005.

\bibitem{li2020gesac}
Jiayuan Li, Qingwu Hu, and Mingyao Ai.
\newblock {GESAC}: Robust graph enhanced sample consensus for point cloud
  registration.
\newblock {\em ISPRS}, 2020.

\bibitem{li2019usip}
Jiaxin Li and Gim~Hee Lee.
\newblock Usip: Unsupervised stable interest point detection from 3d point
  clouds.
\newblock In {\em ICCV}, 2019.

\bibitem{li2020end}
Lei Li, Siyu Zhu, Hongbo Fu, Ping Tan, and Chiew-Lan Tai.
\newblock End-to-end learning local multi-view descriptors for 3d point clouds.
\newblock In {\em CVPR}, 2020.

\bibitem{lowe2004distinctive}
David~G Lowe.
\newblock Distinctive image features from scale-invariant keypoints.
\newblock {\em IJCV}, 2004.

\bibitem{luo2019contextdesc}
Zixin Luo, Tianwei Shen, Lei Zhou, Jiahui Zhang, Yao Yao, Shiwei Li, Tian Fang,
  and Long Quan.
\newblock {ContextDesc}: Local descriptor augmentation with cross-modality
  context.
\newblock In {\em CVPR}, 2019.

\bibitem{luo2018geodesc}
Zixin Luo, Tianwei Shen, Lei Zhou, Siyu Zhu, Runze Zhang, Yao Yao, Tian Fang,
  and Long Quan.
\newblock {GeoDesc}: Learning local descriptors by integrating geometry
  constraints.
\newblock In {\em ECCV}, 2018.

\bibitem{luo2020aslfeat}
Zixin Luo, Lei Zhou, Xuyang Bai, Hongkai Chen, Jiahui Zhang, Yao Yao, Shiwei
  Li, Tian Fang, and Long Quan.
\newblock {ASLFeat}: Learning local features of accurate shape and
  localization.
\newblock In {\em CVPR}, 2020.

\bibitem{mellado2014super}
Nicolas Mellado, Dror Aiger, and Niloy~J Mitra.
\newblock Super 4pcs fast global pointcloud registration via smart indexing.
\newblock In {\em CGF}, 2014.

\bibitem{mises1929praktische}
RV Mises and Hilda Pollaczek-Geiringer.
\newblock Praktische verfahren der gleichungsaufl{\"o}sung.
\newblock {\em ZAMM-Journal of Applied Mathematics and Mechanics/Zeitschrift
  f{\"u}r Angewandte Mathematik und Mechanik}, 1929.

\bibitem{moo2018learning}
Kwang Moo~Yi, Eduard Trulls, Yuki Ono, Vincent Lepetit, Mathieu Salzmann, and
  Pascal Fua.
\newblock Learning to find good correspondences.
\newblock In {\em CVPR}, 2018.

\bibitem{munkres1957algorithms}
James Munkres.
\newblock Algorithms for the assignment and transportation problems.
\newblock {\em Journal of the society for industrial and applied mathematics},
  1957.

\bibitem{myronenko2010point}
Andriy Myronenko and Xubo Song.
\newblock Point set registration: Coherent point drift.
\newblock {\em PAMI}, 2010.

\bibitem{pais20203dregnet}
G~Dias Pais, Srikumar Ramalingam, Venu~Madhav Govindu, Jacinto~C Nascimento,
  Rama Chellappa, and Pedro Miraldo.
\newblock {3DRegNet}: A deep neural network for 3d point registration.
\newblock In {\em CVPR}, 2020.

\bibitem{paszke2017automatic}
Adam Paszke, Sam Gross, Soumith Chintala, Gregory Chanan, Edward Yang, Zachary
  DeVito, Zeming Lin, Alban Desmaison, Luca Antiga, and Adam Lerer.
\newblock Automatic differentiation in pytorch.
\newblock In {\em NeurIPS-W}, 2017.

\bibitem{peng2019pvnet}
Sida Peng, Yuan Liu, Qixing Huang, Xiaowei Zhou, and Hujun Bao.
\newblock {PVNet}: Pixel-wise voting network for 6dof pose estimation.
\newblock In {\em CVPR}, 2019.

\bibitem{perera2012maximal}
Samunda Perera and Nick Barnes.
\newblock Maximal cliques based rigid body motion segmentation with a rgb-d
  camera.
\newblock In {\em ACCV}. Springer, 2012.

\bibitem{poiesi2020distinctive}
Fabio Poiesi and Davide Boscaini.
\newblock Distinctive 3d local deep descriptors.
\newblock {\em arXiv}, 2020.

\bibitem{pomerleau2015review}
Fran{\c{c}}ois Pomerleau, Francis Colas, and Roland Siegwart.
\newblock A review of point cloud registration algorithms for mobile robotics.
\newblock 2015.

\bibitem{qi2017pointnet}
Charles~R Qi, Hao Su, Kaichun Mo, and Leonidas~J Guibas.
\newblock {PointNet}: Deep learning on point sets for 3d classification and
  segmentation.
\newblock In {\em CVPR}, 2017.

\bibitem{quan2020compatibility}
Siwen Quan and Jiaqi Yang.
\newblock Compatibility-guided sampling consensus for 3-d point cloud
  registration.
\newblock {\em TGRS}, 2020.

\bibitem{rodola2013scale}
Emanuele Rodol{\`a}, Andrea Albarelli, Filippo Bergamasco, and Andrea Torsello.
\newblock A scale independent selection process for 3d object recognition in
  cluttered scenes.
\newblock {\em IJCV}, 2013.

\bibitem{rusu2009fast}
Radu~Bogdan Rusu, Nico Blodow, and Michael Beetz.
\newblock Fast point feature histograms (fpfh) for 3d registration.
\newblock In {\em ICRA}, 2009.

\bibitem{sahloul2020accurate}
Hamdi~M Sahloul, Shouhei Shirafuji, and Jun Ota.
\newblock An accurate and efficient voting scheme for a maximally all-inlier 3d
  correspondence set.
\newblock {\em PAMI}, 2020.

\bibitem{sarlin2020superglue}
Paul-Edouard Sarlin, Daniel DeTone, Tomasz Malisiewicz, and Andrew Rabinovich.
\newblock Superglue: Learning feature matching with graph neural networks.
\newblock In {\em CVPR}, 2020.

\bibitem{schops2019bad}
Thomas Schops, Torsten Sattler, and Marc Pollefeys.
\newblock Bad slam: Bundle adjusted direct rgb-d slam.
\newblock In {\em CVPR}, 2019.

\bibitem{shi2020robin}
Jingnan Shi, Heng Yang, and Luca Carlone.
\newblock {ROBIN}: a graph-theoretic approach to reject outliers in robust
  estimation using invariants.
\newblock {\em arXiv}, 2020.

\bibitem{sun2020acne}
Weiwei Sun, Wei Jiang, Eduard Trulls, Andrea Tagliasacchi, and Kwang~Moo Yi.
\newblock {ACNe}: Attentive context normalization for robust
  permutation-equivariant learning.
\newblock In {\em CVPR}, 2020.

\bibitem{wang2018non}
Xiaolong Wang, Ross Girshick, Abhinav Gupta, and Kaiming He.
\newblock Non-local neural networks.
\newblock In {\em CVPR}, 2018.

\bibitem{Wang_2019_ICCV}
Yue Wang and Justin~M. Solomon.
\newblock Deep closest point: Learning representations for point cloud
  registration.
\newblock In {\em ICCV}, 2019.

\bibitem{wang2019prnet}
Yue Wang and Justin~M Solomon.
\newblock {PRNet}: Self-supervised learning for partial-to-partial
  registration.
\newblock In {\em NeuIPS}, 2019.

\bibitem{whelan2015elasticfusion}
Thomas Whelan, Stefan Leutenegger, R Salas-Moreno, Ben Glocker, and Andrew
  Davison.
\newblock {ElasticFusion}: Dense slam without a pose graph.
\newblock Robotics: Science and Systems, 2015.

\bibitem{yang2019polynomial}
Heng Yang and Luca Carlone.
\newblock A polynomial-time solution for robust registration with extreme
  outlier rates.
\newblock {\em arXiv}, 2019.

\bibitem{Yang20arXivTEASER}
Heng Yang, Jingnan Shi, and Luca Carlone.
\newblock {TEASER}: Fast and certifiable point cloud registration.
\newblock {\em arXiv}, 2020.

\bibitem{yang2019performance}
Jiaqi Yang, Ke Xian, Peng Wang, and Yanning Zhang.
\newblock A performance evaluation of correspondence grouping methods for 3d
  rigid data matching.
\newblock {\em PAMI}, 2019.

\bibitem{yang2017performance}
Jiaqi Yang, Ke Xian, Yang Xiao, and Zhiguo Cao.
\newblock Performance evaluation of 3d correspondence grouping algorithms.
\newblock In {\em 3DV}, 2017.

\bibitem{yang2019ranking}
Jiaqi Yang, Yang Xiao, Zhiguo Cao, and Weidong Yang.
\newblock Ranking 3d feature correspondences via consistency voting.
\newblock {\em Pattern Recognition Letters}, 2019.

\bibitem{yang2019extreme}
Zhenpei Yang, Jeffrey~Z Pan, Linjie Luo, Xiaowei Zhou, Kristen Grauman, and
  Qixing Huang.
\newblock Extreme relative pose estimation for rgb-d scans via scene
  completion.
\newblock In {\em CVPR}, 2019.

\bibitem{yew2020RPMNet}
Zi~Jian Yew and Gim~Hee Lee.
\newblock {RPM-Net}: Robust point matching using learned features.
\newblock In {\em CVPR}, 2020.

\bibitem{zeng20173dmatch}
Andy Zeng, Shuran Song, Matthias Nie{\ss}ner, Matthew Fisher, Jianxiong Xiao,
  and Thomas Funkhouser.
\newblock {3DMatch}: Learning local geometric descriptors from rgb-d
  reconstructions.
\newblock In {\em CVPR}, 2017.

\bibitem{zhang2019learning}
Jiahui Zhang, Dawei Sun, Zixin Luo, Anbang Yao, Lei Zhou, Tianwei Shen, Yurong
  Chen, Long Quan, and Hongen Liao.
\newblock {OANet}: Learning two-view correspondences and geometry using
  order-aware network.
\newblock In {\em ICCV}, 2019.

\bibitem{zhao2019nm}
Chen Zhao, Zhiguo Cao, Chi Li, Xin Li, and Jiaqi Yang.
\newblock {NM-Net}: Mining reliable neighbors for robust feature
  correspondences.
\newblock In {\em CVPR}, 2019.

\bibitem{zhao2019robust}
Ruibin Zhao, Mingyong Pang, Caixia Liu, and Yanling Zhang.
\newblock Robust normal estimation for 3d lidar point clouds in urban
  environments.
\newblock {\em Sensors}, 2019.

\bibitem{zhou2018learning}
Lei Zhou, Siyu Zhu, Zixin Luo, Tianwei Shen, Runze Zhang, Mingmin Zhen, Tian
  Fang, and Long Quan.
\newblock Learning and matching multi-view descriptors for registration of
  point clouds.
\newblock In {\em ECCV}, 2018.

\bibitem{zhou2016fast}
Qian-Yi Zhou, Jaesik Park, and Vladlen Koltun.
\newblock Fast global registration.
\newblock In {\em ECCV}, 2016.

\bibitem{zhou2018open3d}
Qian-Yi Zhou, Jaesik Park, and Vladlen Koltun.
\newblock {Open3D}: A modern library for 3d data processing.
\newblock {\em arXiv}, 2018.

\end{thebibliography}
